\documentclass[pdflatex,sn-mathphys-ay,iicol]{sn-jnl}
\usepackage[utf8]{inputenc}
\usepackage[T1]{fontenc}
\usepackage{amsmath,amssymb,amsfonts}
\usepackage{booktabs}
\usepackage{array}
\usepackage{graphicx}
\usepackage{listings}
\usepackage{hyperref}
\hypersetup{hypertexnames=false,colorlinks=true,linkcolor=blue,citecolor=blue,urlcolor=blue}
\usepackage{microtype}
\usepackage{algorithm}
\usepackage{array}
\usepackage{makecell}
\usepackage{algpseudocode}
\usepackage{placeins}
\usepackage{float}
\usepackage{stfloats}
\usepackage{cuted}
\usepackage{xcolor}
\usepackage{soul}
\newcommand{\hiddencomment}[1]{}
\usepackage{rotating} 
\usepackage[most]{tcolorbox}
\usepackage[switch]{lineno}

\newcommand{\R}{\mathbb{R}}

\newcommand{\softplus}{\operatorname{softplus}}
\newcommand{\sigmoid}{\operatorname{sigmoid}}

\newcommand{\clip}{\operatorname{clip}}

\newcolumntype{L}[1]{>{\raggedright\arraybackslash}p{#1}}
\newcolumntype{R}[1]{>{\raggedleft\arraybackslash}p{#1}}
\newcolumntype{S}{>{\centering\arraybackslash}m{0.082\textwidth}}
\newcolumntype{M}{>{\raggedright\arraybackslash}m{0.215\textwidth}}

\definecolor{boxback}{RGB}{255,248,232}
\definecolor{boxrule}{RGB}{204,113,44}
\definecolor{boxtitleback}{RGB}{238,153,74}

\newtcolorbox{postitcard}[1]{%
enhanced,
breakable=false,
colback=boxback,
colframe=boxrule,
colbacktitle=boxtitleback,
coltitle=black,
fonttitle=\bfseries\footnotesize,
title={#1},
arc=2mm,
boxrule=0.7pt,
left=3pt,
right=3pt,
top=3pt,
bottom=2pt,
before skip=3pt,
after skip=5pt
}
\lstdefinestyle{orangebox}{
basicstyle=\ttfamily\scriptsize,
breaklines=true,
frame=single,
backgroundcolor=\color{boxback},
rulecolor=\color{boxrule},
framerule=0.7pt,
xleftmargin=2pt,
xrightmargin=2pt,
aboveskip=2pt,
belowskip=2pt,
showstringspaces=false
}

\title{GUIDER 2.0: Switch-Aware Human Intent Inference for Mobile Manipulation}

\author*[1]{\fnm{Cesar Alan} \sur{Contreras}} \email{cac214@bham.ac.uk} 
\author[2]{\fnm{Manolis} \sur{Chiou}} 
\author[1,3]{\fnm{Alireza} \sur{Rastegarpanah}} 
\author[4]{\fnm{Michal} \sur{Szulik}}
\author*[1]{\fnm{Rustam} \sur{Stolkin}} \email{r.stolkin@bham.ac.uk}

\affil[1]{
    \footnotesize
    \orgdiv{School of Metallurgy and Materials},
    \orgname{University of Birmingham},
    \city{Birmingham},
    \country{United Kingdom}
}
\affil[2]{
    \footnotesize
    \orgdiv{School of Electronic Engineering and Computer Science},
    \orgname{Queen Mary University of London},
    \city{London},
    \country{United Kingdom}
}
\affil[3]{
    \footnotesize
    \orgdiv{School of Computer Science and Digital Technologies},
    \orgname{Aston University},
    \city{Birmingham},
    \country{United Kingdom}
}
\affil[4]{
    \footnotesize
    \orgname{United Kingdom National Nuclear Laboratory},
    \orgdiv{Warrington Professional Hub},
    \orgaddress{\street{5th Floor, Chadwick House, Birchwood Park},
    \city{Warrington},
    \postcode{WA3 6AE},
    \country{United Kingdom}}
\vspace{-1.2em}
}

\AddToHook{cmd/printabstract/before}{\vspace{-2.5em}}
\begin{document}

\abstract{%
Shared autonomy for mobile manipulation requires the robot to infer the operator’s goal during motion. A common challenge arises when the operator switches from one map region, target object, or grasp approach to another mid-trajectory. This paper advances Global User Intent Dual-phase Estimation for Robots (GUIDER) by making it switch-aware. GUIDER 2.0 uses a neural detector to estimate when the operator’s intent changes and incorporates that evidence into its probabilistic belief update.

The same network architecture detects changes in navigation intent (which region the operator is moving toward) and manipulation intent (which object the operator is reaching for). A temporal convolutional network (TCN) and a gated recurrent unit (GRU) encode the previous three seconds of motion. An event-level multilayer perceptron (MLP) combines recent scores and motion summaries to detect intent switches. We generated the training data in simulation. At each decision, a large language model (LLM) selected from navigation targets or grasp options supplied by the simulator and provided a justification based on the episode context. The resulting datasets contain 100,000 mobile-base trajectories and 75,000 manipulator trajectories.

On test data, the detector identifies 96.9\% of navigation intent switches and 99.11\% of manipulation intent switches within 0.5~s of onset. On the original GUIDER dataset, GUIDER 2.0 correctly predicts the final target region in 100/100 trials. Its object and grasping modes each rank the correct target first at contact in 100/100 trials, compared with 99/100 for each original mode. Both pooled paired comparisons show significantly earlier sustained correct predictions. In 65\% of 20 recorded real-robot manipulation episodes, the grasping mode of GUIDER 2.0 reaches a sustained correct prediction earlier than the original grasping mode, with a mean improvement of 1.73~s across all episodes.

\vspace{-1em}
}

\keywords{Intent inference, shared autonomy, mobile manipulation, intent change detection, hazard model, large language models, dataset generation
\vspace{-3em}
}

\maketitle

\begin{figure*}[h]
  \centering
  \includegraphics[width=\textwidth,keepaspectratio]{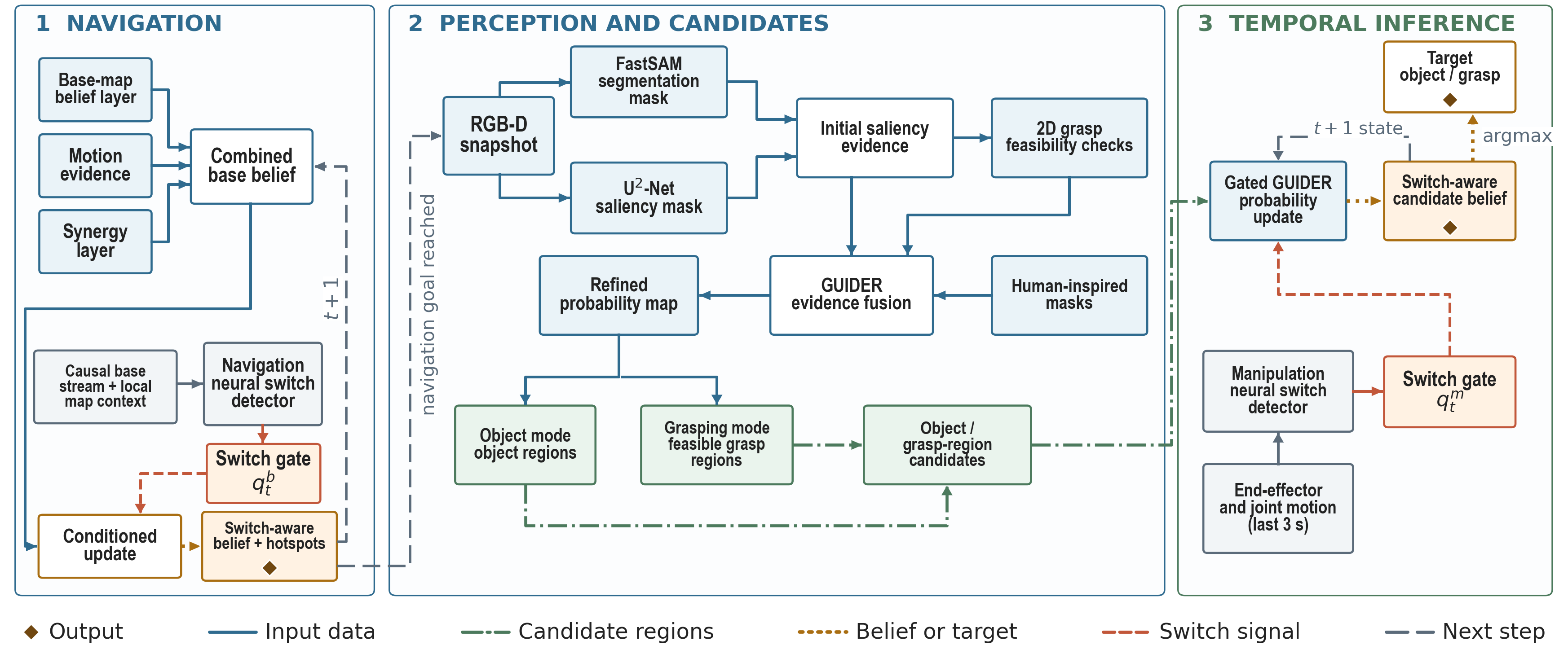}
  \vspace{-2.0em}
  \caption{GUIDER 2.0 at runtime. The neural detector uses only recent motion and, for navigation, local map information to detect intent switches. GUIDER 2.0 uses this evidence to update beliefs over navigation regions, objects, or grasp candidates.}
  \label{fig:guider_integration_flow}
\end{figure*}

\section{Introduction}\label{sec:introduction}
Mobile manipulators operated or supervised by a human are used in situations where the environment is unpredictable and errors can have a very high operational cost. Examples of such applications are remote inspection, hazardous materials handling, nuclear decommissioning and assisting users with reduced mobility. Shared autonomy is used to reduce the operator's workload by combining human commands with robot assistance, provided that the robot’s estimate of the operator’s goal maintains low latency and is stable during normal motion but can respond rapidly when the operator changes the robot's direction.

Our earlier work, the Global User Intent Dual-phase Estimation for Robots (GUIDER) framework \citep{contreras2025guider}, proposed a probabilistic backbone for human intent prediction in mobile manipulation applications, combining observed robot motion with human-inspired spatial priors, geometric action feasibility and mapping context to produce beliefs over navigation and manipulation candidates. The current work builds on that backbone to address a practical challenge for human intent estimators: detecting \textit{mid-trajectory redirection} from motion. An operator may start off moving towards one target and then redirect to a different target when a path is blocked, a grasp is no longer feasible or their task strategy changes. If the intent estimator assumes intent is constant then the posterior beliefs can get "\textit{stuck}" on the first target. If the switch detector is too aggressive then it can result in assistance oscillating between the two candidates in the case of ambiguous motion.

Recent work in shared autonomy has used visual information and language to recognise a wider range of operator goals and adapt assistance as a task progresses \citep{liu2025casper,belsare2025zeroshot,tao2025lams}. Language models support high-level semantic reasoning about why an operator might change goals as the task evolves. Online assistance also depends on detecting when that change occurs, which motivates our use of a compact neural detector that can recognise redirection from robot motion in real time.

Our approach has two stages. For offline training, we used a simulator and a large language model (LLM) to generate labelled trajectories and trained a neural detector on them. During online operation, only the trained detector is needed to estimate from recent robot motion when the operator’s intent changes.

The main idea is to give each simulated redirection a reason grounded in the task. The simulator provides a finite set of navigation targets or feasible object–grasp pairs. Using the episode context, the LLM selects a target that matches the simulator’s requested event kind and explains its choice. Once the motion planner has validated the resulting trajectory, we train the intent switch detector on short windows of robot motion, using changes in the recorded target selections to label when intent switches occur.

This paper introduces GUIDER 2.0, which incorporates switch evidence from a neural detector trained on simulated trajectories generated with a context-conditioned LLM policy (Fig.~\ref{fig:guider_integration_flow}). As evidence of redirection builds, GUIDER 2.0 reduces the influence of the previously favoured target when recent motion supports an alternative. This helps the newly intended navigation region, object or grasp become the leading estimate sooner. GUIDER 2.0 retains the original decision interface and keeps feasibility checks and region constraints outside the neural detector.

\subsection{Contributions}
The paper makes five contributions. \textbf{First}, it defines a bounded LLM policy. The simulator identifies reachable navigation targets or feasible object–grasp pairs, and the LLM uses episode context to choose or switch among them. \textbf{Second}, it proposes an online intent change detector whose encoder combines a temporal convolutional network (TCN) with a gated recurrent unit (GRU). It includes auxiliary training heads and an event-level multilayer perceptron (MLP) postprocessor. \textbf{Third}, it applies the same detector architecture to mobile-base and manipulator motion. \textbf{Fourth}, it incorporates switch evidence into GUIDER’s probabilistic base, object and grasping updates. \textbf{Fifth}, it trains the detector on 100,000 mobile-base and 75,000 manipulation trajectories and evaluates GUIDER 2.0 on the Isaac Sim dataset \citep{contreras2025guider} and recorded real-robot manipulation data \citep{kenny2026guider}.

\section{Related Work}
\subsection{Shared Autonomy and Intent}
Shared autonomy systems often use Bayesian filtering to infer which of several discrete goals the operator intends. The operator’s commands and the robot’s state update the posterior probability of each goal, guiding when to assist and how to arbitrate between human and robot commands \citep{panagopoulos2021bayesian,hoffman2024intentreview,aronson2024adaptation,jonnavittula2024sari,lorthioir2025landmark}. Related work examines how legible motion helps operators infer the robot’s goal and how predictable trajectories help them anticipate its actions \citep{pascher2023motionintent}. Both perspectives matter when intent must be inferred online with low latency.

Recent work has pushed shared autonomy beyond predefined goal sets, allowing robots to infer goals as a task unfolds. CASPER uses vision-language models to infer the operator’s intent and activate the corresponding parameterised assistive skill \citep{liu2025casper}. Vision-only zero-shot inference can also identify targets directly from the observed scene \citep{belsare2025zeroshot}. In teleoperation, LLM-driven mode switching adapts the type of assistance to the task phase and the operator’s behaviour \citep{tao2025lams}. As goals emerge, high-frequency assistance still requires a well-calibrated belief over those goals that remains stable through brief variations in motion and updates fast when the operator redirects.

\subsection{Motion and Goal Switching}
Recurrent models, graph-based predictors and action distribution models can learn patterns in an operator’s trajectory that help infer intent \citep{song2024robottrajectron,chi2023diffusionpolicy}. Even a smooth trajectory, however, can lead towards a new goal. Bayesian online goal change detection and switching state models offer a way to distinguish continued pursuit of the same goal from redirection \citep{tsaknaki2025bocpdar}. In GUIDER 2.0, the TCN-GRU uses motion observed so far to estimate whether the current sample falls within the 0.5~s window following an intent switch. An event-level MLP estimates the event probability. A selected threshold, a debounce and a rising-edge rule convert this probability into switch pulses, while GUIDER 2.0 continues to update its target beliefs from observed motion between intent switches.

Candidate goals can appear in any order and change during a task. As the scene changes, candidate regions may split, merge or disappear. Permutation-invariant networks and Deep Sets encoders score each candidate against a representation of the whole set \citep{kimura2024permutation}. This motivates stable candidate selection in GUIDER’s belief interface regardless of ordering. The switch detector remains goal-agnostic, using motion features to detect a redirection.

\subsection{Goals and Perception}
When navigation goals are unknown in advance, frontier-based exploration generates candidates along the boundary between known free space and unexplored space \citep{popovic2024aipp,sun2025frontiernet}. Samples along the robot’s heading and nodes in a topological map can supplement these frontier goals. GUIDER maintains a belief over a bounded set of map-derived candidates. In GUIDER 2.0, the switch detector uses recent motion to assess whether the operator is redirecting towards a different goal.

For manipulation, targets come from what the robot can see in the scene. Segment Anything and its real-time variants support visual prompting or rapid generation of object proposals \citep{kirillov2023sam,zhao2023fastsam,ravi2024sam2}. Vision-language-action models such as RT-2 and OpenVLA extend the role of perception by using visual and language inputs to predict robot actions at scale \citep{brohan2023rt2,kim2024openvla}. Although these models provide context for assistance, teleoperation still requires confidence gating, feasibility checks, and fail-safe decisions. GUIDER 2.0 preserves the existing GUIDER checks while using the switch detector to supply real-time evidence of intent changes.

\begin{figure*}[!t]
\centering
\includegraphics[width=\textwidth]{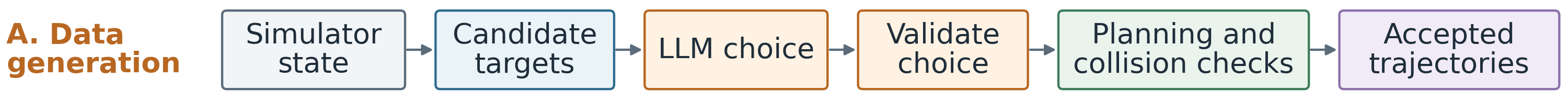}\par
\vspace{-0.40em}
\includegraphics[width=\textwidth]{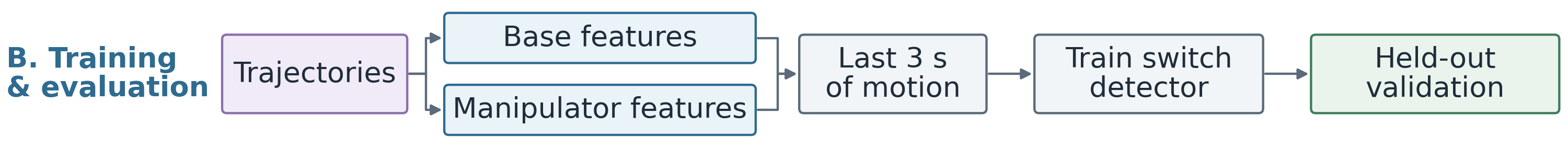}\par
\vspace{-0.40em}
\includegraphics[width=\textwidth]{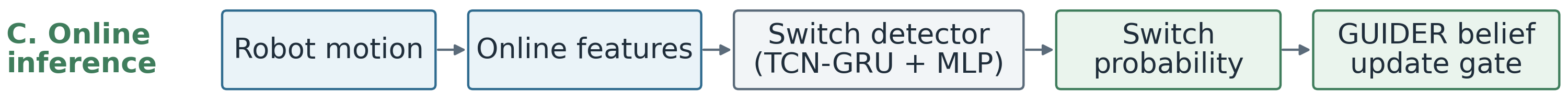}\par
\vspace{-0.40em}
\caption{GUIDER 2.0 uses validated LLM choices to generate trajectories (a), trains and evaluates the switch detector on motion windows (b), and uses switch evidence to update beliefs online (c).}
\label{fig:system_overview}
\end{figure*}

\subsection{Constrained LLM Planning}
LLMs can help select robot actions when their choices are grounded in what the robot can do. SayCan combined language likelihood with estimates of action feasibility, while Inner Monologue used feedback during execution to guide subsequent choices \citep{ahn2022saycan,huang2022inner}. Code as Policies, ProgPrompt, LLM-Planner, LLM+P and Voyager explored how language models can generate action programs, work with planners or build executable skills \citep{liang2022cap,singh2022progprompt,song2023llmplanner,liu2023llmp,wang2023voyager}. More recent work examines how the robot’s state, available actions and feedback can inform planning \citep{li2025multirobot,largemodel2025survey,shaji2026language,xu2026roboagent}.

LMQL and Outlines constrain which responses an LLM can generate \citep{beurerkellner2023lmql,willard2023outlines}. Research on structured generation examines how reliably and efficiently these constraints can be enforced \citep{geng2025jsonschema,park2025grammar,mundler2025diffusion}. Shielding checks proposed actions against explicit rules before execution \citep{carr2022shielding,tappler2022automata}. We apply this principle to dataset generation by accepting the LLM’s choice only if it belongs to the simulator’s current set of possible candidates and matches the requested event kind. The planner then checks the resulting motion for collisions. Figure~\ref{fig:system_overview} gives an overview of data generation, detector training and evaluation, and online inference in GUIDER 2.0.

\section{System Overview}
\begin{figure}[htbp]
  \centering
  \vspace{-1.20em}
  \includegraphics[width=\columnwidth]{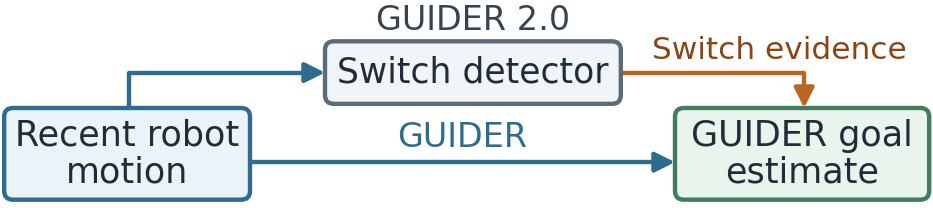}
  \vspace{-1.50em}
  \caption{GUIDER 2.0 uses recent robot motion and switch evidence to update its goal estimate.}
  \vspace{-1.80em}
  \label{fig:intent_overview}
\end{figure}

GUIDER estimates the operator’s intended navigation region, target object or grasp approach using motion evidence, feasibility checks and task constraints \citep{contreras2025guider}. As Fig.~\ref{fig:intent_overview} shows, GUIDER 2.0 incorporates switch evidence into the same belief update while retaining GUIDER’s candidate features and priors.

\subsection{Intent States} \label{sec:Intent-states}
The switch detector uses the last three seconds of robot motion to estimate whether the operator has changed goals. Let $t=1,2,\ldots$ be the estimator index. Navigation updates follow processed odometry and command data, manipulation updates follow end-effector states, and learning and evaluation use a fixed period $\Delta t$.

For \textbf{base motion}, we generate candidate navigation regions online from the occupancy map. Let $N_t^b$ denote the number of candidate regions at time $t$. Then
\vspace{-1em}
\begin{equation*}
\mathcal G_t=\{g_t^{(1)},\ldots,g_t^{(N_t^b)}\},\qquad s_t^b\in\{1,\ldots,N_t^b\}.
\end{equation*}
\vspace{-2em}

$\mathcal{G}_t$ is the bounded set of goals from the map, and state $s_t^b=i$ denotes intent toward $g_t^{(i)}$. The candidate count $N_t^b$ changes as frontiers appear, merge, or disappear.

For \textbf{manipulation}, $N_t^m$ is the number of retained grasp-region candidates at time $t$.
Let $\mathcal O_t^m=\{o_t^{(1)},\ldots,o_t^{(N_t^m)}\}$ be the retained grasp regions and $\mathcal K_{\mathrm{app}}=\{1,\ldots,K_{\mathrm{app}}\}$ the approach classes. Feasible region-approach pairs form $\mathcal C_t^m\subseteq\{1,\ldots,N_t^m\}\times\mathcal K_{\mathrm{app}}$, and
\vspace{-1em}
\begin{equation*}
s_t^m=(i_t,k_t)\in\mathcal C_t^m.
\end{equation*}
\vspace{-2em}

This joint-state representation captures both which feasible graspable region the operator intends to use and how the operator approaches the region.

\subsection{Motion and Context}
At time $t$, the motion vectors are
\vspace{-1em}
\begin{equation*}
x_t^b=[p_t^b,v_t^b,u_t^b],\qquad x_t^m=[p_t^{ee},v_t^{ee},u_t^{ee}],
\end{equation*}
\vspace{-2em}

where base position and velocity are planar, end-effector position and velocity are three-dimensional, and $u_t^d$ is the corresponding command. In the simulator data, each motion candidate has a feature vector $\phi_t^{d,(i)}$. Navigation features include relative distance, command alignment, path cost, clearance, frontier size, and synergy scores. Manipulation features include relative object position, velocity alignment, object size, approach direction, grasp feasibility, reachability, collisions, masked grasp saliency, and target usage. These features describe the kind of scene and task context an operator might consider when choosing a target. The LLM uses selected features to choose among the simulator-provided candidates.
\vspace{-1em}
\begin{equation*}
\Phi_t^d=\{\phi_t^{d,(1)},\ldots,\phi_t^{d,(N_t^d)}\},\qquad d\in\{b,m\}.
\end{equation*}
\vspace{-3em}

\begin{figure*}[!b]
\centering
\includegraphics[width=\textwidth]{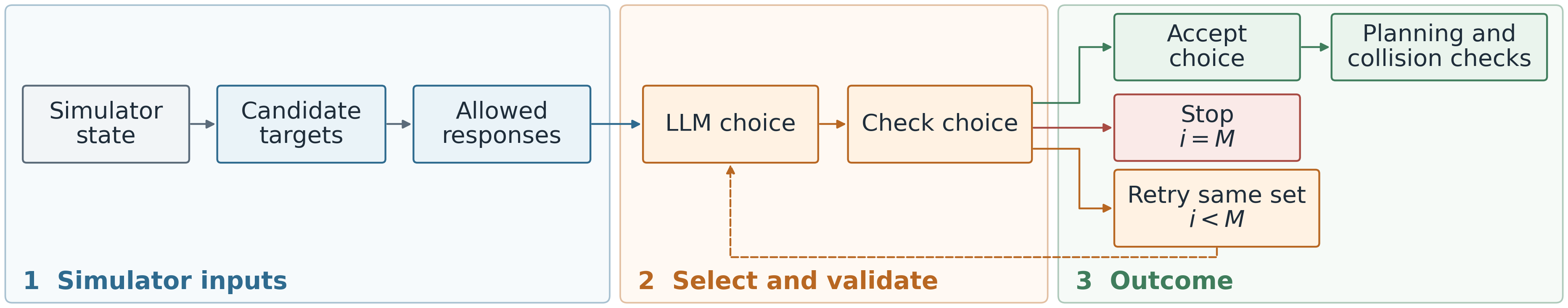}
\vspace{-1.5em}
\caption{LLM choices during dataset generation. Valid choices proceed to planning. An invalid choice triggers a retry with the same candidates until the attempt limit is reached.}
\label{fig:llm_decision_flow}
\end{figure*}

\section{Bounded LLM Policy} \label{}
\subsection{Target Selection}
We generated navigation and manipulation episodes in PyBullet \citep{coumans2016pybullet}. The LLM chose among the available targets using episode context such as current progress, object labels, proximity and target history (Fig.~\ref{fig:llm_decision_flow}). At an intent-change event in domain $d$, let $s_t^d$ denote the current simulated intent state and $\chi_t$ the event kind requested by the episode generator. For that event, the generator constructs a finite, non-empty set of valid candidates
\vspace{-1em}
\begin{equation*}
C_t^d=\{c_1^d,\ldots,c_{n_d}^d\},\qquad n_d>0,\quad d\in\{b,m\}.
\end{equation*}
\vspace{-2em}

Each element $c_j^d$ is a simulator candidate\footnote{These candidates are analogous to the navigation and manipulation intent states in Section~\ref{sec:Intent-states}.}. For navigation, $C_t^b$ contained shelf identifiers or map-region identifiers, depending on the episode type. For manipulation, $C_t^m$ contained target-grasp candidates as pairs $(q,g)$ generated by the simulator, where $q$ denotes an object or handle target and $g$ denotes a valid grasp or approach index.

Let $E$ be the prompt builder, using domain-specific inputs for \(d\in\{b,m\}\). It builds
\vspace{-1em}
\begin{equation*}
x_t^{L}=E(s_t^d,C_t^d,\chi_t),
\end{equation*}
\vspace{-2.2em}

where $x_t^{L}$ contains a scene summary, current target, event kind, target history, object labels, candidate descriptions, candidate positions, target-usage counts, and, when relevant, generator-selected grasp indices.

\subsection{LLM Response Validation}
Let $M$ be the maximum number of allowed language-model responses for one event. For attempt $i\in\{1,\ldots,M\}$, let $\mathcal{J}_t$ be the response schema built from $C_t^d$, and let $\varepsilon_{i-1}^{\mathrm{val}}$ be the previous validation error when a retry is required, with $\varepsilon_0^{\mathrm{val}}$ denoting no prior validation error.
\noindent For a decoded response $a$, let $\nu(a)$ be the selected candidate, let $\kappa(a)$ denote the event kind supplied by the generator, and let $h(a)$ be the natural-language reason (e.g object proximity, grasp feasibility, etc.). The LLM response distribution is $p_\omega$, with parameters $\omega$. The sampled text response is
\vspace{-1em}
\begin{equation*}
\rho_t^{(i)}\sim p_\omega(\cdot\mid x_t^{L},\mathcal{J}_t,\varepsilon_{i-1}^{\mathrm{val}}).
\end{equation*}
\vspace{-2em}

Let $\delta$ decode any response that satisfies the response schema into a decision $a$, let $\bot$ denote rejection, let $G_t^d$ be the domain-specific consistency predicate, and let $H_t^d$ be the full domain validation predicate.
The deterministic validator becomes
\vspace{-1em}
\begin{equation*}
V_t(\rho)=
\begin{cases}
\delta(\rho), & \rho\models\mathcal{J}_t\text{ and }H_t^d(\delta(\rho)),\\
\bot, & \text{otherwise}.
\end{cases}
\end{equation*}
\vspace{-1.4em}

The predicate $H_t^d(a)$ abbreviates $\nu(a)\in C_t^d$, $\kappa(a)=\chi_t$, and $G_t^d(a)$. 

For \textbf{navigation}, $G_t^b(a)$ holds if and only if the decoded candidate is a member of $C_t^b$. For \textbf{manipulation}, $G_t^m(a)$ requires that the decoded target-grasp pair $\nu(a)=(q,g)$ belongs to $C_t^m$ and that the reason $h(a)$ passes the checks for the requested event kind. Let
\vspace{-1em}
\begin{equation*}
\mathcal{A}_t=\{i\in\{1,\ldots,M\}:V_t(\rho_t^{(i)})\neq\bot\}
\end{equation*}
\vspace{-2em}

be the set of attempt numbers for which validation succeeds. The accepted decision and selected candidate are
\vspace{-1em}
\begin{equation*}
(a_t,d_t)=
\begin{cases}
\left(V_t(\rho_t^{(i^\star)}),
\nu(V_t(\rho_t^{(i^\star)}))\right),
& \substack{\mathcal{A}_t\neq\emptyset,\\
i^\star=\min\mathcal{A}_t},
\vspace{0.5em}
\\
(\bot,\bot), & \mathcal{A}_t=\emptyset.
\end{cases}
\end{equation*}
\vspace{-1em}

Here $a_t$ is the accepted decision, and $d_t$ is the selected candidate. If no response satisfies the constraints, the planner receives no candidate for that event. Thus every LLM-selected candidate is either a member of the valid candidate set or is absent:
\vspace{-1em}
\begin{equation*}
\forall t,\quad d_t\in C_t^d\;\lor\;d_t=\bot.
\end{equation*}
\vspace{-2em}

Let $E$ be the prompt builder, $L_\omega$ the LLM text sampler, $V$ the deterministic validator, $R$ the function that converts the selected candidate from the accepted decision into a planner request, and $P$ the planner and collision validator. The executed policy\footnote{The LLM selects a candidate; the motion planner generates the trajectory.} is the composition
\vspace{-1em}
\begin{equation*}
\pi_{\mathrm{exec}}=P\circ R\circ V\circ L_\omega\circ E.
\end{equation*}
\vspace{-2.4em}

\subsection{Context-Guided Selection}
The policy selects candidates based on semantic continuity, manipulability, proximity, and usage history (to encourage a wider spread of targets). The decision log records each accepted choice and the LLM's stated reason. This is intended to reflect how an operator might use task context to change their intended target during redirection.

This constrained role aligns with embodied-agent work in which language models reason over grounded state, memory, affordances, valid actions, feedback, and explanation while a separate robot layer grounds execution \citep{naeem2025sdt,goebel2025classical,shaji2026language}. During data generation, the robot layer is the simulator’s motion planner. During evaluation, GUIDER 2.0 updates intent estimates using neural switch evidence.
  
\subsection{Goal Acceptance and Retries}
Validation distinguishes form rejection (invalid response form), candidate rejection (invalid index or failed domain-specific consistency check), and planning rejection (no reachable collision-free trajectory). The LLM has up to $M$ attempts to select a valid target, which is accepted for execution only if the planner finds a collision-free path to it. The accepted domain sets are $C_t^b$ for shelf or map-region goals and $C_t^m$ for valid target-grasp pairs. Figures~\ref{fig:mobile_llm_contract_cards} and~\ref{fig:manipulator_llm_contract_cards} show the corresponding inputs and accepted outputs.

\begin{figure}[h]
\vspace{-1.2em}
\centering
\small
\begin{postitcard}{Mobile-base LLM input and accepted output}
\scriptsize
\vspace{-0.5em}
\textbf{Input.} The simulator provides the intent state $s_t^b$ (current shelf or map goal), the requested event kind $\chi_t$ (same shelf/new approach, nearby shelf, or diverse shelf/map goal), and the finite set of feasible candidates $C_t^b$. It also supplies object labels, candidate descriptions, positions, usage counts, recent target history and scene context.\\
\textbf{Accepted output.} The target selected by index $\nu(a)\in C_t^b$, the requested event kind $\kappa(a)=\chi_t$, and a short reason $h(a)$ for audit and consistency checks. Waypoints, velocities, paths and robot commands are excluded.
\end{postitcard}
\vspace{-0.6em}
\caption{Mobile-base target selection during dataset generation.}
\label{fig:mobile_llm_contract_cards}
\vspace{-2.0em}
\end{figure}

\begin{figure}[h]
\vspace{-1.2em}
\centering
\small
\begin{postitcard}{Manipulator LLM input and accepted output}
\scriptsize
\vspace{-0.5em}
\textbf{Input.} The simulator provides the intent state $s_t^m$ (current object, task context and support labels), the requested event kind $\chi_t$ (same target/regrasp, nearby/context object, or different/random object), and the finite set of valid target-grasp pairs $C_t^m$. It also supplies object labels, descriptions, positions, target history, usage counts and valid grasp or approach indices.\\
\textbf{Accepted output.} The target-grasp pair selected by index $\nu(a)=(q,g)\in C_t^m$, the requested event kind $\kappa(a)=\chi_t$, and a short reason $h(a)$ for audit and consistency checks. Pose targets, trajectories, newly generated grasps and robot commands are excluded.
\end{postitcard}
\vspace{-0.6em}
\caption{Manipulator grasp selection during dataset generation}
\label{fig:manipulator_llm_contract_cards}
\vspace{-1.6em}
\end{figure}

The LLM response must satisfy the response schema, and the selected candidate must belong to $C_t^d$ and match the requested event kind, with each retry using the same candidate set. Figure~\ref{alg:bounded_llm_generation} shows how candidates are generated, how the LLM response is checked and when the planner tests the selected candidate.

\begin{figure}[h]
\vspace{-1.2em}
\centering
\fbox{%
\begin{minipage}{0.96\columnwidth}
\begin{enumerate}
\setlength{\itemsep}{0pt}\setlength{\parsep}{0pt}\setlength{\topsep}{2pt}\setlength{\partopsep}{0pt}
\scriptsize
\item Given $d\in\{b,m\}$, $s_t^d$, $\chi_t$ and attempt limit $M$, build the finite set $C_t^d=\{c_1^d,\ldots,c_{n_d}^d\}$, $n_d>0$. Its candidates are shelf or map-region goals for $d=b$ and valid target-grasp pairs $(q,g)$ for $d=m$.
\item Build $x_t^L=E(s_t^d,C_t^d,\chi_t)$ using scene context, current target, event kind, target history, candidate descriptions, positions, usage counts and, where relevant, valid grasp indices.
\item Use $C_t^d$ to build the response schema $\mathcal{J}_t$. Set $\varepsilon_0^{\mathrm{val}}\gets\emptyset$.
\item For $i=1,\ldots,M$, sample $\rho_t^{(i)}\sim p_\omega(\cdot\mid x_t^L,\mathcal{J}_t,\varepsilon_{i-1}^{\mathrm{val}})$ and decode $a=(\nu(a),\kappa(a),h(a))$. Accept only if $\rho_t^{(i)}\models\mathcal{J}_t$, $\nu(a)\in C_t^d$, $\kappa(a)=\chi_t$ and $G_t^d(a)$.
\item On acceptance, set $a_t\gets a$ and $d_t\gets\nu(a_t)$. Return $d_t$ if it passes the planner’s reachability and collision checks. Otherwise, return $\bot$.
\item On validation failure, record the error in $\varepsilon_i^{\mathrm{val}}$ and retry. If all attempts fail, return $\bot$.
\end{enumerate}
\end{minipage}}
\caption{Bounded LLM selection and motion planning checks.}
\label{alg:bounded_llm_generation}
\vspace{-0.0em}
\end{figure}

\FloatBarrier

\begin{figure*}[!hb]
\centering
\includegraphics[width=\textwidth]{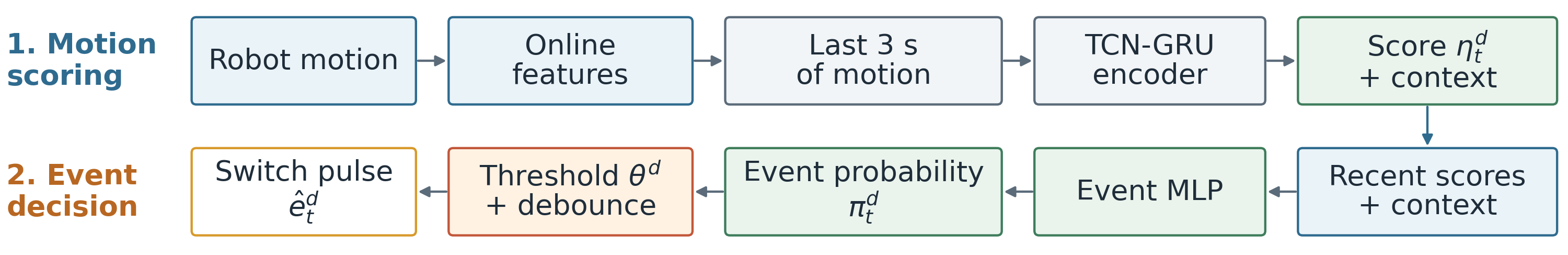}
\vspace{-1.80em}
\caption{The TCN-GRU scores recent motion. The MLP combines recent scores and motion summaries into an event probability. A rising edge in the debounced threshold decision produces a switch pulse for GUIDER 2.0.
}
\label{fig:final_detector_flow}
\end{figure*}

\section{Intent Switch Detector} \label{sec:final_switch_model}
\subsection{Detector Interface}
The detector consumes only features available by the current motion sample and during inference everything else (candidate identities, LLM outputs, labels, event times, target poses, and future state) is excluded. The reported detector comprises the TCN-GRU core, auxiliary-loss training, event-level MLP, which are evaluated together in the reported experiments.
The final detector has two learned stages. The first stage is a TCN-GRU that scores the current sample from a window covering the previous three seconds of motion. The second stage is a second-stage MLP that converts recent core scores and motion summaries into an event probability. A validation-selected threshold and fixed 0.05\,s debounce define an active state whose rising edge produces a switch pulse for GUIDER 2.0. Figure \ref{fig:final_detector_flow} shows the two-stage switch detector workflow.

\subsection{Intent Switch Labels}
The task is online intent switch detection from robot trajectory streams. At estimator row $t$, the model may use only information available at row $t$ or earlier. It must decide whether an operator redirection event is occurring for either domain $d\in\{b,m\}$.
The domain-specific builder converts each trajectory into a feature vector built from current and earlier samples $X_t^d$ for row $t$ and domain $d$.
The evaluated detector detects changes of destination or target object. Let $\iota_t^d$ be the stable target identity used for switch labels in domain $d$ at row $t$. For the mobile base, $\iota_t^b$ is the shelf or map-goal destination regardless of changes in approach to that destination. For manipulation, $\iota_t^m$ is the object or handle target after projecting away same-target grasp or approach retries. For $t>1$, the raw target-redirection indicator is
\vspace{-1em}
\begin{equation*}
e_t^d=\mathbb{I}[\iota_t^d\neq\iota_{t-1}^d].
\end{equation*}
\vspace{-2em}

The first row has no previous row and is treated as a non-event.
Events that change the destination or object become positive training events. Same-shelf approach changes and same-target regrasp or retry events remain in the simulated trajectories, but they are negative for this evaluated target-switch detector because the underlying target did not change.

Let $\Delta t=0.05$ s be the row sampling period, and $W=9$ so that each event contributes the ten sampled offsets $0,\ldots,9$ used by the training builders for the 0.5 s detection target. For indices before the first row, define $e_s^d=0$ when $s<1$. The supervised target marks the event row and the following nine sampled rows:
\vspace{-1em}
\begin{equation*}
y_t^d=\max_{j=0,\ldots,W} e_{t-j}^d.
\end{equation*}
\vspace{-1.6em}

This defines the sampled 0.5 s detection target used for training, so $y_t^d=1$ means row $t$ lies in the positive post-switch training window. 

\subsection{Domain-Specific Inputs}
The shared architecture uses domain-specific input features and trained weights. The base input table uses a history of planar motion sampled over three seconds expressed relative to the current base frame: body-frame displacement, yaw change, body velocity, angular velocity, commanded velocity, commanded angular velocity, acceleration, angular acceleration, speed, a local map, and motion summaries.

The manipulator input table uses the same sampling pattern over the previous three seconds, but with end-effector and joint data (end-effector displacement, current-axis alignment, end-effector velocity, speed, acceleration, and the manipulator joint positions and velocities). Thus, the manipulator detector learns redirection from motion consistency and changes in the learned motion representation.

\subsection{Motion Encoding and Scoring}
Let $R$ be the number of sampled history lags, and let $L_r$ be the $r$th lag, measured in rows, for $r=1,\ldots,R$. Let $\xi_{t-L_r}^d$ be the normalized feature token from row $t-L_r$ in domain $d$, and let $c_t^d$ be the normalized non-temporal context vector available at row $t$. The input to the neural core\footnote{This compact sequence is used so the network can compare recent motion with older motion while using only observations available at the current sample.} is
\vspace{-1em}
\begin{equation*}
U_t^d=[\xi_{t-L_1}^d,\ldots,\xi_{t-L_R}^d,c_t^d].
\end{equation*}
\vspace{-2em}

Samples span the previous three seconds and are sampled more densely near the current time. Let $\Theta$ denote the learned encoder and hazard-head parameters. A temporal convolutional front end and GRU encode the sequence as
\vspace{-1em}
\begin{equation*}
z_t^d=f_{\Theta}(U_t^d)\in\R^{192},
\end{equation*}
\vspace{-2em}

where $f_{\Theta}$ is the learned encoder and $z_t^d$ is the 192-dimensional latent state. The latent state is useful because the postprocessor can later measure changes in the same representation used by the hazard head. The hazard head maps the latent state to a probability:
\vspace{-1em}
\begin{equation*}
\ell_t^d=g_{\Theta}(z_t^d),
\qquad
\eta_t^d=\sigmoid(\ell_t^d).
\end{equation*}
\vspace{-2em}

Here $g_{\Theta}$ is the learned hazard head\footnote{This hazard is an intent-change score, not a collision or physical-danger score.}, $\ell_t^d$ is its logit, and $\eta_t^d$ is the predicted probability that row $t$ lies in the post-switch detection window.

\subsection{Training Objective}
Both domains use the same objective family. Let $w_1$ be the positive-row class weight and $w_0$ be the negative-row class weight; $w_{y_t^d}$ selects the appropriate weight for row $t$. The main loss is weighted binary cross-entropy (BCE)
\vspace{-1em}
\begin{equation*}
\mathcal{L}_{\mathrm{haz}}^d =
-\sum_t w_{y_t^d}
\left[y_t^d\log \eta_t^d+(1-y_t^d)\log(1-\eta_t^d)\right]
\end{equation*}
\vspace{-1.5em}

This loss trains the core score to be high inside the post-switch window and low during stable motion. 

Auxiliary heads shape the latent representation without adding runtime information. Let $g_{\mathrm{prox}}$, $g_{\mathrm{onset}}$, and $g_{\mathrm{fore}}$ be learned heads for proximity, onset, and future-motion prediction. For a raw event at row $\tau$, the event-proximity target is a linear post-event decay over the same post-switch label window used by $y_t^d$:
\vspace{-1em}
\begin{equation*}
r_t^d=\max_{\tau:e_\tau^d=1}\mathbb{I}[0\le t-\tau\le W]\frac{W+1-(t-\tau)}{W+1}.
\end{equation*}
\vspace{-1.5em}

If no event row satisfies the condition, the maximum is zero. This target is positive only on rows that are also positive under $y_t^d$ and is largest at the raw switch row. Let $\hat r_t^d$ be the network's proximity prediction, and let $\hat e_t^d$ be the onset prediction for the raw event row $e_t^d$. The auxiliary predictions and losses are:
\vspace{-1em}
\begin{align*}
& \hspace{4.5em} \hat r_t^d=\sigmoid(g_{\mathrm{prox}}(z_t^d)),\\
& \hspace{5em} \mathcal{L}_{\mathrm{prox}}^d=\sum_t (r_t^d-\hat r_t^d)^2,\\
& \hspace{4.5em} \hat e_t^d=\sigmoid(g_{\mathrm{onset}}(z_t^d)),\\
& \mathcal{L}_{\mathrm{onset}}^d=-\sum_t\left[e_t^d\log\hat e_t^d
+(1-e_t^d)\log(1-\hat e_t^d)\right]
\end{align*}
\vspace{-1.5em}

The proximity head teaches closeness to an event, and the onset head trains the encoder to identify the switch sample. Let $H=\{5,10\}$ be the supervised future horizons in rows. For the mobile base, $m_t^b$ concatenates, for each $h\in H$, the body-frame displacement, $\sin$ of yaw change, and speed change between rows $t$ and $t+h$. For the manipulator, $m_t^m$ concatenates, for each $h\in H$, end-effector displacement in world coordinates and end-effector speed change between rows $t$ and $t+h$. Rows without enough future samples are masked out of this auxiliary term.
Before computing the Euclidean future-motion loss, each component is normalized by a positive component scale. Let $S_{\mathrm{fore}}^d$ be the diagonal matrix containing those scales for domain $d$, in the same order as the components of $m_t^d$. Let $\hat m_t^d=g_{\mathrm{fore}}(z_t^d)$ be the dimensional prediction. The normalized target and prediction are
\vspace{-0.6em}
\begin{equation*}
\bar{m}_t^d=(S_{\mathrm{fore}}^d)^{-1}m_t^d,\qquad
\hat{\bar{m}}_t^d=(S_{\mathrm{fore}}^d)^{-1}\hat m_t^d.
\label{eq:future_motion_normalization}
\end{equation*}
\vspace{-1em}

The future-motion loss over valid future-label rows $\mathcal{V}^d$ is
\vspace{-1em}
\begin{equation*}
\mathcal{L}_{\mathrm{fore}}^d=\sum_{t\in\mathcal{V}^d}\|\bar{m}_t^d-\hat{\bar{m}}_t^d\|_2^2.
\end{equation*}
\vspace{-1.5em}

This term helps the encoder learn motion change structure, but $m_t^d$ is not available to the runtime detector.

Let $\lambda_{\mathrm{haz}}$, $\lambda_{\mathrm{prox}}$, $\lambda_{\mathrm{onset}}$, and $\lambda_{\mathrm{fore}}$ be non-negative loss weights. The full objective is
\vspace{-1em}
\begin{align*}
\mathcal{L}^d = \\
&\hspace{-1.5em} \lambda_{\mathrm{haz}}\mathcal{L}_{\mathrm{haz}}^d
+\lambda_{\mathrm{prox}}\mathcal{L}_{\mathrm{prox}}^d + \lambda_{\mathrm{onset}}\mathcal{L}_{\mathrm{onset}}^d
+\lambda_{\mathrm{fore}}\mathcal{L}_{\mathrm{fore}}^d
\end{align*}
\vspace{-2em}

The final runs used $\lambda_{\mathrm{haz}}=4.0$, $\lambda_{\mathrm{prox}}=3.0$,
$\lambda_{\mathrm{onset}}=2.0$, $\lambda_{\mathrm{fore}}=0.1$, AdamW decoupled weight decay $10^{-5}$, chunk length 64 and stride 8.

\subsection{Event-Level Switch Detection}
Training followed the same order in both domains (base and manipulation). The second detector stage is a small MLP over current and earlier score features. Its input includes the current hazard-head score, lagged probabilities at one, two, and five rows, short-window maxima and means, probability deltas, and the latent-state shift between recent windows.

The mobile-base postprocessor adds speed, speed change, turn mean, command-turn change, and command-speed change. The manipulator postprocessor adds end-effector speed, end-effector speed change, end-effector axis change, joint-speed mean and maximum, joint-motion path length, and joint-acceleration mean.

The postprocessor is a one-hidden-layer MLP with 64 hidden units and Gaussian error linear unit (GELU) activations trained for five epochs with AdamW learning rate $10^{-3}$ and weight decay $10^{-4}$. Let $a_t^d$ be the postprocessor logit, $\pi_t^d=\sigmoid(a_t^d)$ be its event probability, and let $y_t^d$ be the same post-event target used for core training. The row BCE term is
\vspace{-1em}
\begin{equation*}
\mathcal{L}_{\mathrm{row}}^d=
-\sum_t\left[y_t^d\log\pi_t^d+(1-y_t^d)\log(1-\pi_t^d)\right].
\end{equation*}
\vspace{-2em}

Let $\tau_q^d$ be the raw event row for true event $q$ in domain $d$.
The positive row interval for that event is
\begin{equation*}
\mathcal{R}_q^d=\{t:\tau_q^d\le t\le \tau_q^d+W\}.
\end{equation*}
The early-event interval $\mathcal{E}_q^d=\mathcal{R}_q^d$ because the reporting target is detection within 0.5 s.

Let $\mathcal{N}_{\mathrm{stable}}^d$ be stable negative rows, and let $\mathcal{N}_{\mathrm{same}}^d$ be same-target or same-destination negative rows. The event-level term encourages at least one high logit inside each event interval:
\vspace{-1em}
\begin{equation*}
\mathcal{L}_{\mathrm{bag}}^d=\sum_q
\softplus\left(1-\max_{t\in\mathcal{R}_q^d}a_t^d\right).
\end{equation*}
\vspace{-1em}

The early-event term applies the same loss to $\mathcal{E}_q^d$:
\vspace{-1em}
\begin{equation*}
\mathcal{L}_{\mathrm{early}}^d=\sum_q
\softplus\left(1-\max_{t\in\mathcal{E}_q^d}a_t^d\right).
\end{equation*}
\vspace{-1em}

For a non-empty negative row set $\mathcal{N}$, let $\operatorname{Top}_{12}(\mathcal{N})$ return the rows with the twelve largest logits, or all rows if fewer than twelve are available. The negative top-logit penalty is
\vspace{-1em}
\begin{equation*}
\mathcal{L}_{\mathrm{top}}^d(\mathcal{N})=
\frac{1}{|\operatorname{Top}_{12}(\mathcal{N})|}
\sum_{t\in\operatorname{Top}_{12}(\mathcal{N})}\softplus(a_t^d+0.5).
\end{equation*}
\vspace{-1.5em}

If a batch contains no rows from $\mathcal{N}_{\mathrm{stable}}^d$ or $\mathcal{N}_{\mathrm{same}}^d$, the corresponding top-logit term is omitted for that batch.

Let $\alpha_{\mathrm{same}}^b=0.25$ and $\alpha_{\mathrm{same}}^m=0.0$. The complete postprocessor loss is
\vspace{-1em}
\begin{align*}
\mathcal{L}_{\mathrm{pp}}^d
&=0.2\mathcal{L}_{\mathrm{row}}^d
+2.5\mathcal{L}_{\mathrm{bag}}^d
+1.5\mathcal{L}_{\mathrm{early}}^d \nonumber\\
&\quad+0.05\mathcal{L}_{\mathrm{top}}^d(\mathcal{N}_{\mathrm{stable}}^d)
+\alpha_{\mathrm{same}}^d\mathcal{L}_{\mathrm{top}}^d(\mathcal{N}_{\mathrm{same}}^d).
\end{align*}
\vspace{-2em}

Since $\mathcal{E}_q^d=\mathcal{R}_q^d$ in the evaluated configuration, the bag and early-event losses are equivalent and contribute a combined weight of 4.0. The final validation-selected postprocessor thresholds were 0.76 for the mobile base and 0.55 for the manipulator. Let $\theta^b=0.76$, $\theta^m=0.55$, and $\delta_{\mathrm{deb}}=0.05$ s. The raw postprocessor decision is $u_t^d=\mathbb{I}[\pi_t^d\ge\theta^d]$. Let $\rho_t^d=\min\{\rho\le t\mid u_j^d=1\ \forall j\in[\rho,t]\}$ be the first row of the current active run when $u_t^d=1$, and let $T_t$ be the row timestamp. The debounced active state is
\vspace{-0.6em}
\begin{equation*}
b_t^d=
\begin{cases}
\mathbb{I}[T_t-T_{\rho_t^d}\ge\delta_{\mathrm{deb}}], & u_t^d=1,\\
0, & u_t^d=0.
\end{cases}
\end{equation*}
\vspace{-1em}

With $b_0^d=0$, the switch pulse is the rising edge of this debounced state:
\vspace{-1em}
\begin{equation*}
\hat e_{\mathrm{alarm},t}^d=b_t^d(1-b_{t-1}^d).
\end{equation*}
\vspace{-2em}

If a subthreshold row occurs, the active run is reset and $\rho_t^d$ is undefined until the next active row, so debouncing suppresses isolated one-row spikes. Because $\hat e_{\mathrm{alarm},t}^d$ is a pulse rather than an active-state indicator, one sustained threshold crossing contributes at most one runtime alarm. 

\section{GUIDER 2.0 Integration} \label{sec:guider_integration_method}
GUIDER is our probabilistic intent-estimation framework for mobile manipulation \citep{contreras2025guider} that operates in two phases. In the \textbf{mobile-base phase}, it maintains belief over likely interaction areas, and in the \textbf{manipulation phase}, it maintains belief over candidate manipulation targets. 

With the trained neural network, we can preserve the original GUIDER interface and add a switch signal that alters the temporal response of the belief update following sustained evidence of \textit{intent redirection}. The event-level MLP estimates the switch probability, which GUIDER 2.0 then converts into a switch gate and applies it within the existing base or manipulation updates.

The integration applies to both parts of GUIDER. For the mobile base, the switch gate modifies update triggering, prediction horizons, decay factors, motion-evidence gain, synergy-map growth, and stale-region suppression. For manipulation, the switch gate modifies probability growth, probability decay, and drop allowance for object or grasp-region candidates. In \textit{object mode}, the retained candidates are object-level regions. In \textit{grasping mode}, the retained candidates are feasible grasp-region candidates produced from the object-support map and the grasp-feasibility masks \citep{kenny2026guider}. The two modes share the same temporal probability-evolution rule but differ in the set of candidate regions to which that rule applies.

\subsection{GUIDER Base} \label{subsec:guider_base_method}

\subsubsection{\textbf{Original GUIDER base}}
The original base update follows \citet{contreras2025guider}. It is restated here, in the same order as the implementation, to define every quantity changed by GUIDER 2.0. Let $\Omega$ be the occupancy-grid cell set, $p(c)=[x,y]^\top$ the world coordinate of cell $c$, and $B_{\mathrm{base}}$, $B_{\mathrm{motion}}$, and $B_{\mathrm{syn}}$ the context, motion-evidence, and synergy layers. The context layer is initialized from occupancy, with $B_{\mathrm{base}}(c,0)=0.02$ for free cells, $0$ for unknown cells, and $1$ for occupied cells.

Non-boundary occupied clusters are inflated into nearby free cells. With $\Omega_{\mathrm{free}}=\{c\in\Omega\mid\mathrm{Occ}(c)=0\}$, define the family of object clusters $\mathfrak{C}_{\mathrm{obj}}$, and normalized cell distance $d_{\mathrm{cell}}(c,C)$ (from free cell $c$ to occupied cluster $C\in\mathfrak{C}_{\mathrm{obj}}$). The inflated context contribution can be written as
\vspace{-1em}
\begin{align*}
\hspace{-0.4em} B_{\mathrm{base}}(c,0)\\ 
&\hspace{-3.0em} \leftarrow \max\!\left(B_{\mathrm{base}}(c,0),\right.\left.0.2+0.4[1-d_{\mathrm{cell}}(c,C)]\right),\\
&\hspace{-1.5em} c\in\Omega_{\mathrm{free}},\quad C\in\mathfrak{C}_{\mathrm{obj}}, \quad d_{\mathrm{cell}}(c,C)\le 1.
\end{align*}
\vspace{-2em}

\noindent And $\mathcal{O}_{\mathrm{static}}\subseteq\Omega$ is the static object-context support used by both original and integrated updates.
\vspace{-1em}
\begin{equation*}
\mathcal{O}_{\mathrm{static}}=
\{c\in\Omega_{\mathrm{free}}\mid \exists C\in\mathfrak{C}_{\mathrm{obj}}\ \text{with}\ d_{\mathrm{cell}}(c,C)\le 1\}
\end{equation*}
\vspace{-1em}

\noindent For planar state $p_t^b=[x_t,y_t]^\top$ and $v_t^b=[v_{x,t},v_{y,t}]^\top$, original GUIDER projects motion over a multi-horizon set $\mathcal{T}_{\mathrm{orig}}=\{5,10,30\}\,\mathrm{s}$ and for horizon $\tau\in\mathcal{T}_{\mathrm{orig}}$, the projected position is $p_{t,\tau}^b=p_t^b+\tau v_t^b$.

Let $d_{\max}>0$ be the implementation radius of the radial motion-evidence kernel. The radial motion evidence at a point sampled along a projected path is
\vspace{-1em}
\begin{equation*}
E_{\tau,t}(c)= \max\left(0, 1-\frac{\|p(c)-p_{t,\tau}^b\|_2}{d_{\max}} \right),
\end{equation*}
\vspace{-1.4em}

and the multi-horizon evidence map is
\vspace{-1em}
\begin{equation*}
\tilde{E}_t(c) = \max_{\tau\in\mathcal{T}_{\mathrm{orig}}}E_{\tau,t}(c).
\end{equation*}
\vspace{-1.4em}

\noindent At each belief update, each layer is multiplied by a retention factor, $\gamma_{\mathrm{base,free}}=0.55$, $\gamma_{\mathrm{base,obj}}=0.50$, $\gamma_{\mathrm{motion}}=0.25$, and $\gamma_{\mathrm{syn}}=0.75$. Thus, the base layer is retained with factor $\gamma_{\mathrm{base,free}}$ for free cells outside $\mathcal{O}_{\mathrm{static}}$ and $\gamma_{\mathrm{base,obj}}$ for object-context cells,
where object-context values are clipped to a minimum of $0.1$, while the motion and synergy layers are retained with factors $\gamma_{\mathrm{motion}}$ and $\gamma_{\mathrm{syn}}$, respectively.

At an update, the motion layer is increased by the motion-evidence field:
\vspace{-0.8em}
\begin{equation*}
\begin{aligned}
B_{\mathrm{motion},t}(c)
&=\clip\!\left(B_{\mathrm{motion},t}^{\mathrm{decay}}(c)
+0.40\tilde{E}_t(c),0,1\right)\\
\end{aligned}
\end{equation*}
\vspace{-1.5em}

The synergy layer activates near static object-context cells intersected by recent motion evidence. Let $r_{\mathrm{syn}}=0.15$ m be the nominal local interaction radius, $\eta_0=0.70$ the reset value, $\eta_{\mathrm{lb}}=0.60$ the lower-bound trigger, $\eta_{\mathrm{inc}}=0.05$ the increment, and $\eta_{\mathrm{cap}}=0.95$ the upper bound on synergy growth.

The visited synergy set, $\mathcal{V}_{\mathrm{syn},t}$, comprises cells for which the static object-context set $\mathcal{O}_{\mathrm{static}}$ is non-empty, recent motion evidence is present, $\tilde{E}_t(c)>0$, and the Euclidean distance to the nearest supported object cell does not exceed $r_{\mathrm{syn}}$. For cells in $\mathcal{V}_{\mathrm{syn},t}$, a decayed synergy value below $\eta_{\mathrm{lb}}$ is reset to $\eta_0$, whereas a value at or above $\eta_{\mathrm{lb}}$ is increased by $\eta_{\mathrm{inc}}$ and capped at $\eta_{\mathrm{cap}}$. Cells outside $\mathcal{V}_{\mathrm{syn},t}$ retain their decayed synergy value. The combined area-belief map, which combines all three layers, is then defined pointwise as the maximum of $B_{\mathrm{base}}$, $B_{\mathrm{motion}}$, and $B_{\mathrm{syn}}$.

\subsubsection{GUIDER 2.0 Base Update}
The GUIDER 2.0 base method retains the original GUIDER map structure and update logic while using the neural detector to condition its temporal response.

Let $\pi_t^b$ be the event probability estimated by the navigation MLP, $\theta^b$ the validation-selected base threshold, and $\varepsilon_\pi=10^{-4}$ the clipping floor. The probability is clipped to $\tilde{\pi}_t^b=\min(1-\varepsilon_\pi,\max(\varepsilon_\pi,\pi_t^b))$. Using the logit $\ell(z)=\log[z/(1-z)]$, the threshold-relative switch drive is $\mu_t^b=\ell(\tilde{\pi}_t^b)-\ell(\theta^b)$. Let $\zeta_t^b$ denote the continuous switch evidence, initialized as $\zeta_0^b=0$. With update interval $\Delta t$, leak time constant $\tau_s=0.35$ s, and evidence gain $\kappa_s=1.0\,\mathrm{s}^{-1}$,
\vspace{-1.0em}
\begin{equation*}
\zeta_t^b = \max\left(0, \zeta_{t-1}^b \exp\left(-\frac{\Delta t}{\tau_s}\right) + \kappa_s\Delta t\,\mu_t^b \right).
\end{equation*}
\vspace{-1.2em}

\noindent The normalized base switch gate is
\begin{equation*}
q_t^b =
\frac{\sigma(s_s(\zeta_t^b-\zeta_s^0))
-\sigma(-s_s\zeta_s^0)}
{1-\sigma(-s_s\zeta_s^0)},
\end{equation*}
where $\sigma(x)=(1+\exp[-x])^{-1}$, $s_s=8.0$, and $\zeta_s^0=0.25$. Thus, $q_t^b=0$ when $\zeta_t^b=0$.

The debounced binary alarm reports runtime switch events, while the continuous gate $q_t^b$ controls GUIDER 2.0's belief evolution. The base implementation normally updates after the robot has moved more than $30$ cm; during active motion, the continuous switch gate additionally permits an update when $q_t^b>0.2$ and $\|v_t^b\|_2>0.02\,\mathrm{m\,s^{-1}}$.

\noindent The prediction horizon set is also switch-conditioned:
\vspace{-1em}
\begin{equation*}
\mathcal{T}_b(q_t^b)=
\begin{cases}
\{2,5,10\}\;\mathrm{s}, & q_t^b>0.3,\\
\{5,10,30\}\;\mathrm{s}, & \text{otherwise}.
\end{cases}
\end{equation*}
\vspace{-1em}

The integrated motion-evidence map is then the maximum evidence across these switch-conditioned horizons, $\tilde{E}_t^{\mathrm{int}}(c;q_t^b)=\max_{\tau\in\mathcal{T}_b(q_t^b)}E_{\tau,t}(c)$.

The gate linearly interpolates the retention factors $(\gamma_{\mathrm{free}},\gamma_{\mathrm{obj}},\gamma_{\mathrm{motion}},\gamma_{\mathrm{syn}})$ from $(0.55,0.50,0.25,0.75)$ at $q_t^b=0$ to $(0.45,0.45,0.08,0.35)$ at $q_t^b=1$. The motion-evidence gain is simultaneously increased according to $\lambda(q_t^b)=0.40(1+2q_t^b)$.

These switch-conditioned factors are applied using the same layer-decay rules as the original base update. Known free cells outside $\mathcal{O}_{\mathrm{static}}$ use $\gamma_{\mathrm{free}}(q_t^b)$, while object-context cells in $\mathcal{O}_{\mathrm{static}}$ use $\gamma_{\mathrm{obj}}(q_t^b)$ with the original $0.1$ floor retained; the motion and synergy layers use $\gamma_{\mathrm{motion}}(q_t^b)$ and $\gamma_{\mathrm{syn}}(q_t^b)$ respectively. The original clipped motion update is otherwise unchanged, replacing the fixed gain $0.40$ with $\lambda(q_t^b)$ and the original multi-horizon evidence $\tilde{E}_t$ with $\tilde{E}_t^{\mathrm{int}}$.

\noindent The continuous switch gate also adjusts the synergy parameters as $r_{\mathrm{syn}}(q_t^b)=(0.15+0.10q_t^b)\,\mathrm{m}$, $\eta_0(q_t^b)=0.70+0.15q_t^b$, and $\eta_{\mathrm{inc}}(q_t^b)=0.05+0.05q_t^b$, while the lower-bound trigger $\eta_{\mathrm{lb}}=0.60$ and cap $\eta_{\mathrm{cap}}=0.95$ remain unchanged.

The original visited-synergy and synergy-update rules are retained with these switch-conditioned quantities. Specifically, a cell is visited when $\mathcal{O}_{\mathrm{static}}$ is non-empty, $\tilde{E}_t^{\mathrm{int}}(c;q_t^b)>0$, and its distance to the nearest static object-context cell does not exceed $r_{\mathrm{syn}}(q_t^b)$. For visited cells, a decayed synergy value below $\eta_{\mathrm{lb}}$ is reset to $\eta_0(q_t^b)$; otherwise it is increased by $\eta_{\mathrm{inc}}(q_t^b)$ up to $\eta_{\mathrm{cap}}$. Unvisited cells retain their decayed synergy value.

The GUIDER 2.0 base method additionally suppresses stale belief when the previously dominant region is poorly supported by current motion evidence. Four-connected components are extracted from cells satisfying $B_{\mathrm{combined}}^{\mathrm{int}}(c,t^-)\ge0.75$, and the component with the greatest summed combined belief is selected as $R_{t^-}^{\mathrm{top}}$. Its current mean motion evidence is $\bar{E}_t(R_{t^-}^{\mathrm{top}})=|R_{t^-}^{\mathrm{top}}|^{-1}\sum_{c\in R_{t^-}^{\mathrm{top}}}\tilde{E}_t^{\mathrm{int}}(c;q_t^b)$. When $q_t^b>0.2$ and $\bar{E}_t(R_{t^-}^{\mathrm{top}})<0.05$, the motion and synergy beliefs within this region are multiplied by $(1-0.7q_t^b)$ and $(1-0.5q_t^b)$ respectively. If no component exists above the $0.75$ threshold, stale-region suppression is skipped.

After stale-region suppression, the integrated combined map is recomputed as the same pointwise maximum of the base, motion, and synergy layers used by the original method.

With non-empty semantic seeds, motion evidence changes belief intensity without connecting separated interaction regions. With $\tau_{\mathrm{hot}}=\tau_{\mathrm{comb}}=0.85$, four-connected components are first extracted from cells for which either $B_{\mathrm{base},t}(c)\ge0.85$ or $B_{\mathrm{syn},t}(c)\ge0.85$. If this semantic seed is empty, the method falls back to four-connected components of cells satisfying $B_{\mathrm{combined}}^{\mathrm{int}}(c,t)\ge0.85$. If both sets are empty, no hotspot is emitted. For each resulting component, the representative hotspot is the component cell with minimum Euclidean distance to its mean spatial position.

The resulting base estimator preserves GUIDER's original map interface while being designed to improve responsiveness during frequent switching.

\subsection{GUIDER Manipulation} \label{subsec:guider_manip_method}

\subsubsection{\textbf{Object-mode perception}}
The original manipulation update follows \citet{contreras2025guider}, with object and grasp-region support from \citet{kenny2026guider}. The complete baseline is retained here in compact form because GUIDER 2.0 changes its temporal evolution but not its perception or feasibility stages.

At manipulation trigger time $t_0$, let $\mathcal{D}_{t_0}=\{I_{t_0},D_{t_0},q_{t_0},T_{t_0}\}$ denote the input data, comprising the RGB image, aligned depth image, end-effector state, and frame transforms available to GUIDER, respectively. The static manipulation-perception pipeline $\mathcal{F}_{\mathrm{static}}(\mathcal{D}_{t_0})$ returns the active candidate set $\hat{\mathcal{O}}_{t_0}=\{o_k\}_{k=1}^{N_{t_0}}$, where $N_{t_0}=|\hat{\mathcal{O}}_{t_0}|$, together with the initial probability vector $\hat{\mathbf{p}}_{t_0}=[p_1(t_0),\ldots,p_{N_{t_0}}(t_0)]^\top$. In object mode, $o_k$ denotes an object candidate; in grasping mode, it denotes a feasible grasp-region candidate.

Let $I:\Omega\subset\mathbb{Z}^2\rightarrow\mathbb{R}^3$ and $D:\Omega\subset\mathbb{Z}^2\rightarrow\mathbb{R}_{\ge0}$ denote the RGB and aligned depth images. For each valid pixel $(u,v)$ with $D(u,v)>0$, the depth is reprojected using the camera intrinsics as 
\begin{equation*}
X(u,v) =
\begin{bmatrix}
\frac{(u-c_x)D(u,v)}{f_x}\\
\frac{(v-c_y)D(u,v)}{f_y}\\
D(u,v)
\end{bmatrix}.
\end{equation*}

\noindent The resulting point cloud is restricted to the near-field manipulation range $0.2\le X_z\le1.2$ m. A dominant support plane $\mathbf{n}^{\top}X+d=0$ is then estimated, and points close to this plane, according to the point-to-plane distance $|\mathbf{n}^{\top}X+d|/\|\mathbf{n}\|_2$, are removed before hierarchical density-based spatial clustering of applications with noise (HDBSCAN). HDBSCAN uses position and surface-normal features 

$\mathbf{f}_i=[x_i,y_i,z_i,\alpha n_x,\alpha n_y,\alpha n_z]$ with $\alpha=0.1$.

Cluster centroids are projected into the image and used as FastSAM prompts. In parallel, U$^2$-Net gives a saliency field $S(u,v)\in[0,1]$. The FastSAM and U$^2$-Net outputs are fused into an initial image-evidence map $P(u,v)$. Pixels supported by one detector receive intermediate evidence, while pixels supported by both receive stronger evidence.

GUIDER then applies a cascading series of feasibility checks comprising bounding-box grasp feasibility, morphological graspability, object-level advanced graspability, and rectangle-level advanced graspability. Starting from $P^{(0)}(u,v)=P(u,v)$, each binary feasibility mask $M_k$ scales supported pixels by $\lambda_{\mathrm{in}}=1.2$ and unsupported pixels by $\lambda_{\mathrm{out}}=0.4$, producing the final mask-scaled field $P^{(K_{\mathrm{mask}})}$ after $K_{\mathrm{mask}}$ masks.

\noindent The mask-scaled evidence is further weighted by image position and depth. The image-center weight is
\vspace{-1em}
\begin{equation*}
w_c(u,v) = \exp\left(-\frac{(u-u_0)^2+(v-v_0)^2} {2\sigma_c^2} \right).
\label{eq:center_bias}
\end{equation*}
\vspace{-1.2em}

\noindent with principal point $(u_0,v_0)$ and $\sigma_c=240$ px, while the depth weight is
\vspace{-1em}
\begin{equation*}
w_d(u,v) = \exp[-\alpha_z(D(u,v)-z_0)^2].
\end{equation*}
\vspace{-2.0em}

\noindent around the nominal manipulation distance $z_0$ with $\alpha_z=1.5\,\mathrm{m}^{-2}$ and $z_0=1.0$ m. Their product defines the weighted evidence field
\vspace{-0.7em}
\begin{equation*}
P_{\mathrm{w}}(u,v)=P^{(K_{\mathrm{mask}})}(u,v)\,w_c(u,v)\,w_d(u,v).
\end{equation*}
\vspace{-2em}

\noindent Positive weighted evidence is floored at $\epsilon=0.01$, while zero-valued pixels remain zero, producing $P^\ast$. Connected components of this final object support define $\mathcal{C}_t^{\mathrm{obj}}=\{\mathcal{O}_i\}_{i=1}^{N_t^{\mathrm{obj}}}$, where $N_t^{\mathrm{obj}}$ is the number of components. Each component receives the mean evidence score $g_i=|\mathcal{O}_i|^{-1}\sum_{(u,v)\in\mathcal{O}_i}P^\ast(u,v)$, which is clipped to the initial temporal probability $p_i(t_0)=\clip(g_i,0,0.99)$.

\subsubsection{\textbf{Grasping-mode perception}}
Grasping mode changes only the support used for candidate extraction. Rather than retaining the complete object-support map, it restricts the final probability support to regions satisfying the advanced graspability rectangle test, giving the operator more localised assistance within an object. 
\noindent Let $A_{\mathrm{rect}}(u,v)$ denote the advanced graspability rectangle mask, obtained by checking object geometry against the two-finger gripper's 85 mm maximum opening and 8.25 mm finger thickness, including clearance at candidate grasp locations. Morphological closing produces the stabilized grasp-region mask $G_{\mathrm{grasp}}(u,v)=\operatorname{close}(A_{\mathrm{rect}}(u,v))$ and the resulting grasping-mode support is $P_{\mathrm{grasp}}(u,v)=P^\ast(u,v)G_{\mathrm{grasp}}(u,v)$. Object mode fully retains object-level candidates, whereas grasping mode retains feasible grasp-region candidates. Candidate scores in both modes are clipped to $[0,0.99]$ before applying the same temporal probability-evolution rule.

\begin{figure}[h]
  \centering
  \includegraphics[width=\columnwidth]{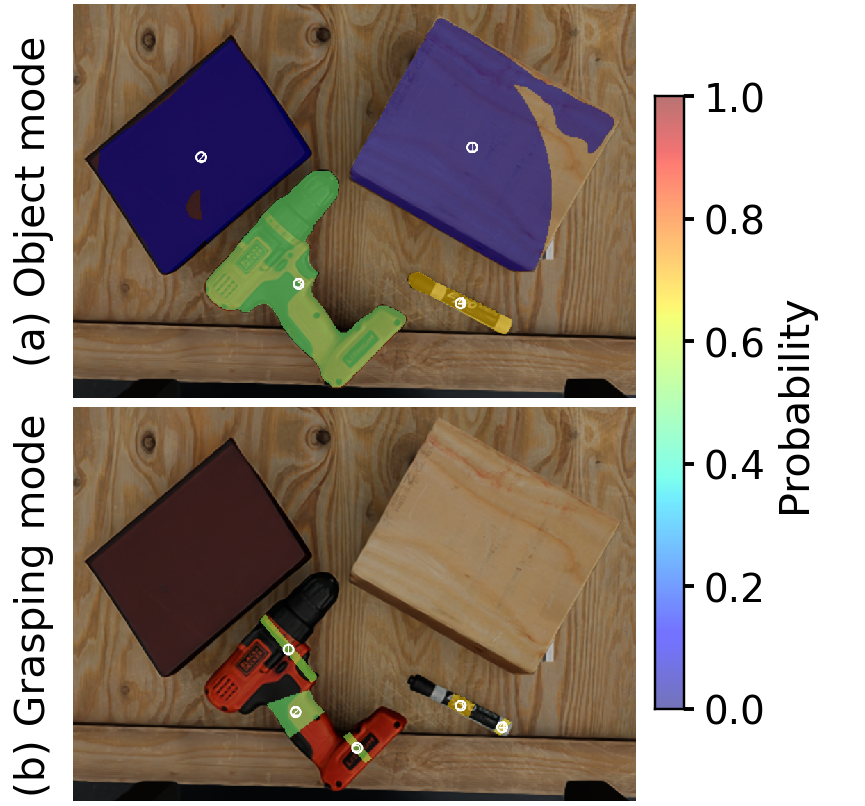}
  \vspace{-1.5em}
  \caption{Object mode tracks probabilities over objects. Grasping mode first limits candidates to feasible grasp regions, then updates their probabilities over time.}
  \label{fig:manip_object_grasping_modes}
  \vspace{-1.8em}
\end{figure}

Figure \ref{fig:manip_object_grasping_modes} shows the difference between object and grasping modes.

\subsubsection{Target Probability Evolution from End-Effector Motion}
Let candidate $i$ have centroid $\mathbf{c}_i=[x_i,y_i,z_i]^\top$ and let the tool-center point be $\mathbf{q}(t)=[x_e(t),y_e(t),z_e(t)]^\top$. GUIDER approximates the two-finger gripper operating region as an upright cylinder of radius $r_g=0.085/3$ m and height $h_g=0.10$ m. The horizontal distance to its axis is $\rho_i(t)=[(x_i-x_e(t))^2+(y_i-y_e(t))^2]^{1/2}$, and its vertical extent is $[z_e(t)-h_g,z_e(t)]$. The corresponding vertical clearance is $h_i(t)=\max(0,z_e(t)-h_g-z_i,z_i-z_e(t))$, giving the candidate-to-cylinder distance $d_i(t)=[\max(0,\rho_i(t)-r_g)^2+h_i(t)^2]^{1/2}$.

A candidate is considered to be approached when its distance decreases relative to the previous retained observation or when $d_i(t)<0.005$ m; denote this indicator by $a_i(t)$ and its complement by $\bar{a}_i(t)=1-a_i(t)$. For a newly introduced candidate with no previously retained counterpart, $d_i(t-\Delta t)$ is initialized to $d_i(t)$, making the distance-decrease condition false on its first observation. End-effector speed and acceleration magnitudes are $v_q(t)=\|\dot{\mathbf{q}}(t)\|_2$ and $a_q(t)=\|\ddot{\mathbf{q}}(t)\|_2$.

The nominal growth term is
\vspace{-1em}
\begin{equation*}
G_i(t) = \frac{\alpha_g}{d_i(t)+\delta} (1+\gamma_v v_q(t)+\gamma_a a_q(t)) (1-0.7\bar{a}_i(t)).
\end{equation*}
\vspace{-1.4em}

The nominal decay term is
\vspace{-1em}
\begin{equation*}
\begin{aligned}
D_i(t) = \alpha_d(d_i(t)+\delta) (1+0.3\gamma_v v_q(t) +\\[0.3ex]
&\hspace{-5em}0.3\gamma_a a_q(t)) (1+0.5\bar{a}_i(t)).
\end{aligned}
\end{equation*}
\vspace{-1.4em}

\noindent Where $\delta=0.1$ m, $\alpha_g=0.08$ m s$^{-1}$, $\alpha_d=0.8$ m$^{-1}$ s$^{-1}$,
$\gamma_v=0.1$ s m$^{-1}$, and $\gamma_a=0.05$ s$^2$ m$^{-1}$. With these units, $G_i(t)$ and $D_i(t)$ have their probability-change rates in s$^{-1}$ under the forward-Euler update. If a candidate starts from a low initial probability, the implementation uses a reduced growth factor for that candidate until it leaves the low-start region.

A short-horizon prediction $\mathbf{q}_{\mathrm{pred}}(t)=\mathbf{q}(t)+T_{\mathrm{pred}}\dot{\mathbf{q}}(t)+\tfrac{1}{2}T_{\mathrm{pred}}^2\ddot{\mathbf{q}}(t)$, with $T_{\mathrm{pred}}=0.3$ s, is used to identify candidates protected from rapid probability reduction.

Let $\mathcal{I}_t$ be the retained candidate-index set after perception and feasibility filtering. In object mode, each $i\in\mathcal{I}_t$ represents an object-level region; in grasping mode, it represents a feasible grasp-region candidate identified by its region and approach label $(r_i,k_i)$. If $\mathcal{I}_t=\emptyset$, no target prediction or probability update is produced. Otherwise, candidate identities are transported from the previous retained set $\mathcal{I}_{t^-}$ using persistent object labels in object mode and grasp-region overlap together with the approach label in grasping mode. Denote this correspondence by $\varphi_t(i)$, with $\varphi_t(i)=\bot$ when no previous counterpart exists. The carried probability $p_i^{-}(t)$ is the previous probability of the matched candidate when $\varphi_t(i)\neq\bot$, and otherwise the bounded initial probability $p_i^0(t)$ assigned by the perception stage. The integrated state uses the analogous carried value $p_i^{\mathrm{int},-}(t)$.

The protected set $\mathcal{T}_{K_{\mathrm{near}}}(t)$ contains the $\min(K_{\mathrm{near}},|\mathcal{I}_t|)$ candidates with the smallest predicted gripper distances at $\mathbf{q}_{\mathrm{pred}}(t)$, together with the currently nearest candidate; candidate index order breaks distance ties. Here $K_{\mathrm{near}}=3$, distinct from the number $K_{\mathrm{app}}$ of grasp approach classes. Protected candidates use the per-update drop allowance $\beta_i(t)=0.002$, while all remaining candidates use $\beta_i(t)=0.005$.
\\

The forward-Euler probability update is
\vspace{-1em}
\begin{align*}
\hspace{3em} \tilde{p}_i(t) = p_i^{-}(t) + \Delta t\left(G_i(t)-D_i(t)\right), 
\\ \qquad i\in\mathcal{I}_t,
\\
p_i(t) = \clip\left(\max[\tilde{p}_i(t),p_i^{-}(t)-\beta_i(t)], 0, p_{\max} \right),
\\ \qquad i\in\mathcal{I}_t,
\end{align*}
\vspace{-2em}

with $p_{\max}=0.99$.

\subsubsection{GUIDER 2.0 Manipulation Update}
The manipulation switch gate is constructed identically to the base gate using the event probability estimated by the manipulation MLP $\pi_t^m$ and validation-selected threshold $\theta^m$. Specifically, $\pi_t^m$ is clipped with the same $\varepsilon_\pi=10^{-4}$ floor to obtain $\tilde{\pi}_t^m$, and the threshold-relative drive is $\mu_t^m=\ell(\tilde{\pi}_t^m)-\ell(\theta^m)$. The leaky evidence state $\zeta_t^m$, initialized as $\zeta_0^m=0$, follows the same recurrence as the base state with $\tau_s=0.35$ s and $\kappa_s=1.0\,\mathrm{s}^{-1}$. The manipulation gate $q_t^m$ uses the same normalized formulation as the base gate, replacing $\zeta_t^b$ with $\zeta_t^m$. Thus, $q_t^m=0$ when $\zeta_t^m=0$.

For each retained candidate $i\in\mathcal{I}_t$, the GUIDER 2.0 manipulation method computes a motion-consistent alternative support score. Let $d_i^{\mathrm{pred}}(t)$ be the gripper distance from candidate $i$ to the predicted end-effector point $\mathbf{q}_{\mathrm{pred}}(t)$. With $\sigma_{\mathrm{alt}}=0.25$ m, define
\vspace{-1em}
\begin{equation*}
\tilde{s}_i(t) = \exp\left(-\frac{d_i^{\mathrm{pred}}(t)}{\sigma_{\mathrm{alt}}} \right)
\begin{cases}
1, & i\in\mathcal{T}_{K_{\mathrm{near}}}(t),\\
0.2, & i\notin\mathcal{T}_{K_{\mathrm{near}}}(t).
\end{cases}
\end{equation*}
\vspace{-1.4em}

The normalized alternative support is
\vspace{-1em}
\begin{equation*}
s_i(t) = \frac{\tilde{s}_i(t)} {\max_{j\in\mathcal{I}_t}\tilde{s}_j(t)+\epsilon_s},
\qquad \epsilon_s=10^{-6}.
\label{eq:manip_alt_support}
\end{equation*}
\vspace{-1.4em}

When a previous retained set exists, let $j_{t^-}^{\mathrm{top,int}}$ be the smallest-index candidate attaining the maximum previous integrated probability. A current candidate is treated as the stale previous top only when it is matched to the previously top-ranked candidate, $\varphi_t(i)=j_{t^-}^{\mathrm{top,int}}$. For that candidate, the mismatch is $m_i(t)=1-s_i(t)$; for all other candidates, or when no previous retained set exists, $m_i(t)=0$.

\noindent The integrated update modifies the original growth, decay, and drop allowance:
\begin{align*}
G_i^{\mathrm{int}}(t) &= G_i(t) \left(1+g_{\mathrm{alt}}q_t^m s_i(t) \right),
\\
D_i^{\mathrm{int}}(t) &= D_i(t) \left( 1+g_{\mathrm{old}}q_t^m m_i(t) \right),
\\
\beta_i^{\mathrm{int}}(t) &= \beta_i(t) \left(1+g_{\mathrm{drop}}q_t^m m_i(t)\right).
\end{align*}

with evaluated gains $g_{\mathrm{alt}}=2.0$, $g_{\mathrm{old}}=3.0$, and $g_{\mathrm{drop}}=4.0$. The integrated probability recurrence retains the original forward-Euler and drop-limited update, using the carried integrated probability $p_i^{\mathrm{int},-}(t)$ and replacing $G_i(t)$, $D_i(t)$, and $\beta_i(t)$ with $G_i^{\mathrm{int}}(t)$, $D_i^{\mathrm{int}}(t)$, and $\beta_i^{\mathrm{int}}(t)$, respectively, while retaining $p_{\max}=0.99$.

Thus, sustained neural switch evidence increases the probability-growth term for the motion-consistent alternatives and increases decay and permits a larger probability drop when motion no longer supports the previous leader. The same switch-aware temporal update is used in both object and grasping modes.

\section{Dataset Generation} \label{sec:dataset_generation_evidence}
The mobile-base and manipulator datasets were generated with similar simulation pipelines. In both domains, the simulator first constructed a finite set of valid intent candidates, then had a local LLM \texttt{Llama-3.1-Nemotron-Nano-4B-v1.1} select a candidate from that set. The LLM was run with TensorRT-LLM using an NVIDIA 4-bit floating-point (NVFP4) engine, an 8192-token runtime context, deterministic decoding, maximum outputs of 32 tokens for mobile-base decisions and 64 tokens for manipulator decisions, and at most six attempts. After candidate selection, the existing planner and PyBullet \citep{coumans2016pybullet} simulator enforced reachability, collision, and manipulation constraints before executing the trajectory in simulation. Table \ref{tab:generation_randomization} summarizes the final generation settings.

\begin{table*}[t]
\centering
\footnotesize
\caption{Simulation settings for the mobile-base and manipulator datasets.}
\label{tab:generation_randomization}
\begin{tabular}{@{}>{\centering\arraybackslash}m{10pt}@{\hspace{4pt}}m{\dimexpr\linewidth-14pt\relax}@{}}
\toprule
\rotatebox[origin=c]{90}{\textbf{Mobile base}} &
\textbf{Scene.} Store bounds $[-12.5,12.5]\times[-7.5,7.5]$~m with $0.10$~m occupancy resolution. Each map has $6$--$16$ border features and $16$--$24$ shelves before reachability filtering.\par
\textbf{Dimensions.} Border-feature depth follows $\mathcal{U}[0.6,2.2]$~m and length follows $\mathcal{U}[0.8,3.0]$~m. Shelf dimensions are $x\sim\mathcal{U}[0.9,2.4]$~m and $y\sim\mathcal{U}[0.25,0.65]$~m, with random long-axis orientation.\par
\textbf{Poses and clearance.} Robot radius is $0.50$~m, with planning and sampling margins of $0.08$~m each. After map inflation, start and goal poses are sampled from valid free-space or shelf-approach candidates. Maps require at least 12 reachable target shelves and a connected free-space fraction $\ge0.92$.\par
\textbf{Intent switch events.} $66\%$ of trajectories contain events, with at most 6 per trajectory. Event progress is $s\in[0,0.98]$, with late-bias exponent $\alpha=3.0$. Of all events, $27\%$ use uniform timing.\par
\textbf{Event weights.} Same shelf with a new approach ($0.315$), nearby shelf in the cone ($0.337$), or random shelf ($0.348$).\par
\textbf{Motion.} Kinematic execution at $\Delta t=0.05$~s, with at most 2400 rows per trajectory. \\
\midrule
\rotatebox[origin=c]{90}{\textbf{Manipulator}} &
\textbf{Scene.} One table, three open cabinets and one closed-handle cabinet, with global rotation $\psi\sim\mathcal{U}[-\pi,\pi]$. Four supports hold 16 YCB objects, with $3$--$5$ per support.\par
\textbf{Object placement.} On the table, $x\sim\mathcal{U}[-0.075,0.225]$~m and $y\sim\mathcal{U}[-0.205,0.205]$~m. On shelves, $x\sim\mathcal{U}[0.065,0.155]$~m and $y\sim\mathcal{U}[-0.230,0.230]$~m.\par
\textbf{Pose perturbations.} Table and open-cabinet anchors vary by $\pm0.01$~m in position and $\pm0.10$~rad in yaw. Before global rotation, the closed cabinet has $x=0.95\pm0.03$~m, $y=\pm0.03$~m and yaw perturbation $\pm0.10$~rad. Object yaw follows $\mathcal{U}[-\pi,\pi]$. Table objects also have roll and pitch sampled from $\mathcal{U}[-0.10,0.10]$~rad.\par
\textbf{Reachability.} Reject objects outside Franka reach, overlapping other objects, penetrating supports or failing target-reachability checks. Shelf objects must stay within cabinet bounds and satisfy cabinet-top clearance.\par
\textbf{Intent switch events.} $66\%$ of trajectories contain events, with at most 4 per trajectory. Event progress is $s\in[0.25,0.92]$, with late-bias exponent $\alpha=2.0$.\par
\textbf{Event weights.} Same target with a new grasp ($0.171$), nearby/context object ($0.329$), or different/random object ($0.500$).\par
\textbf{Motion and planning.} PyBullet execution uses FCL collision checks and OMPL RRTConnect, with three planning attempts and nominal planning time $0.35$~s. The joint-velocity scale is sampled from $[0.55,1.0]$ of the joint limits, with per-joint jitter $[0.90,1.10]$. Maximum joint acceleration follows $\mathcal{U}[10,20]$. \\
\bottomrule
\end{tabular}
\end{table*}

\subsection{Navigation Episodes}
Each mobile-base episode began with an unknown occupancy grid. A simulated range sensor revealed cells as the base moved, after which the generator extracted reachable shelf-approach candidates and formed the finite candidate set $\mathcal{G}_t$. The initial destination and any later redirection target were selected only from this set. Some trajectories included redirection events that changed the destination or its approach at one or more randomized event times and regenerated the remaining path. Execution was kinematic at $\Delta t=0.05$ s.

\begin{figure}[h]
\centering
\includegraphics[width=0.99\columnwidth]{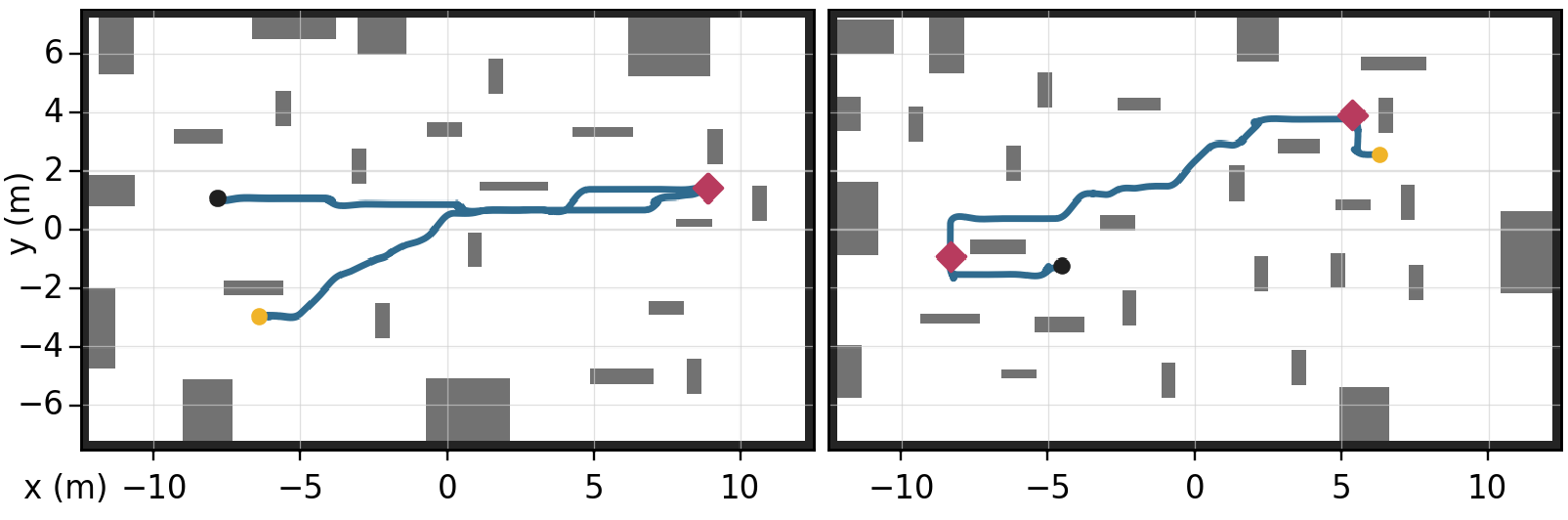}
\vspace{-1.6em}
\caption{Mobile-base trajectories, where black circles mark starts, yellow circles mark final states, and red diamonds mark intent-change locations.}
\label{fig:base-training_trajectory_examples}
\end{figure}

Domain randomization included the placement of obstacles, obstacle density, sensor range, field of view, range noise, map update latency, command noise, and operator style. The saved data included base pose, velocity, command, obstacle distances, candidate features, target information, path cost, event timestamps, and candidate-set identities. Figure \ref{fig:base-training_trajectory_examples} shows an example of generated trajectories, marking situations where redirection events occurred.

\vspace{-1.8em}

\subsection{Manipulation Episodes}
Manipulator episodes used a Franka Panda with a two-finger gripper in PyBullet. Objects were randomized over table and cabinet supports, and each object was associated with a finite set of candidate approach directions. Candidate pre-grasp states were screened by inverse kinematics, reachability, support-clearance, and collision tests before entering the set of valid target-grasp pairs.

\begin{figure}[h]
\vspace{-1.4em}
\centering
\includegraphics[width=0.84\columnwidth]{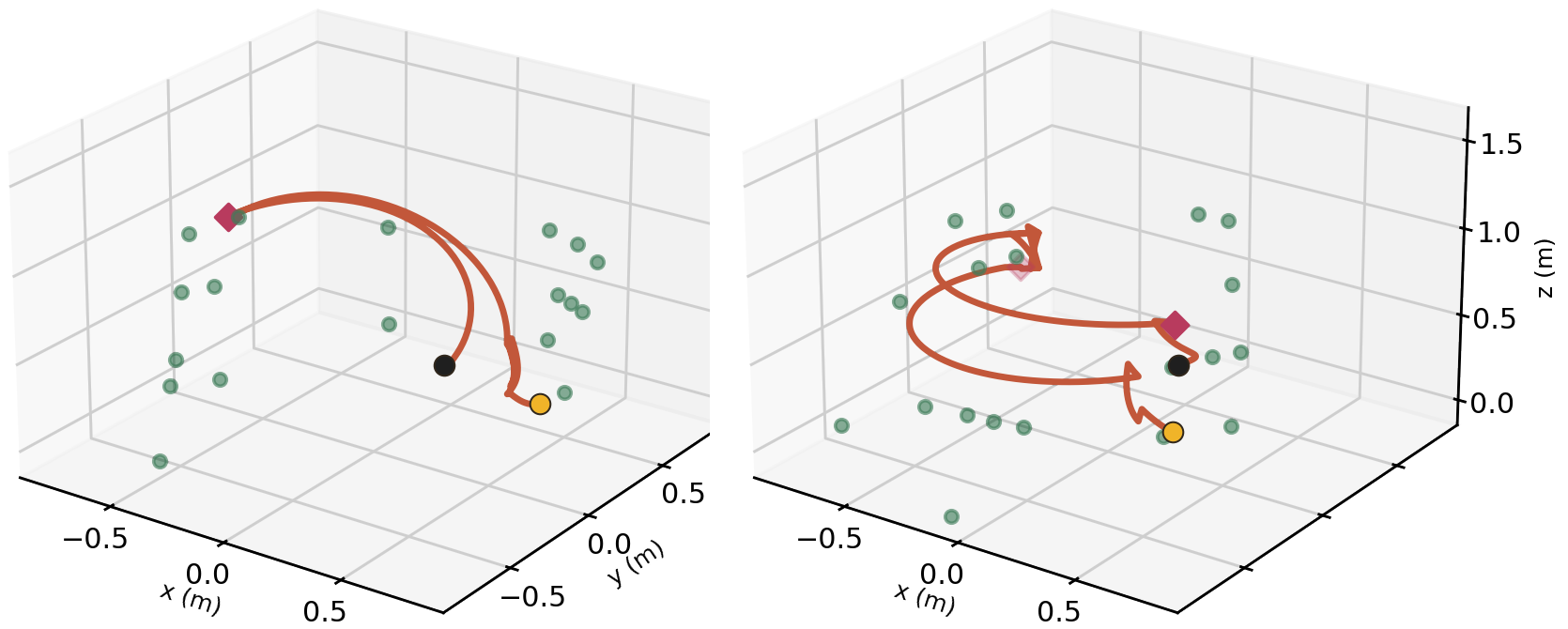}
\caption{Manipulator trajectories. Green circles mark action regions, red diamonds mark intent switches, and black and yellow circles mark starts and ends, respectively.}
\label{fig:manip_training_trajectory_examples}
\vspace{-1.8em}
\end{figure}

Some trajectories included redirection events that changed the target or approach in $(i_t,k_t)$ during execution, triggering replanning from the current robot state. FCL collision checking and OMPL RRTConnect were used with up to three attempts and a $0.35$ s planning limit; the language model selected only feasible target-grasp pairs, while the planner generated and validated the corresponding motion. The dataset retained LLM justification, robot motion, object and support geometry, candidate identities and feasibility, selected target-approach identity, and event timestamps. Figure \ref{fig:manip_training_trajectory_examples} shows example trajectories from the manipulation dataset.

\subsection{Trajectory Composition} \label{subsec:dataset_composition}
The generator produced 100,000 mobile-base trajectories and 75,000 manipulator trajectories, all generated in PyBullet. Table~\ref{tab:current_dataset_composition} gives the generated trajectory composition. Eventful trajectories contain target or approach changes, whereas trajectories without redirection events retain their selected targets.

\begin{table}[h]
\centering
\vspace{-1.6em}
\caption{Generated trajectory composition.}
\label{tab:current_dataset_composition}
\footnotesize
\begin{tabular*}{\columnwidth}{@{\extracolsep{\fill}}lrr@{}}
\toprule
Quantity & Mobile base & Manipulator \\
\midrule
Trajectories               & 100,000 & 75,000 \\
With redirection events    & 66,000  & 49,500 \\
Without redirection events & 34,000  & 25,500 \\
\bottomrule
\end{tabular*}
\vspace{-2.4em}
\end{table}

Mobile-base redirection events are grouped as same shelf/new approach, nearby shelf cone, and random shelf. Manipulator redirection events are grouped as same target/regrasp, nearby/context object, and different/random object. These redirection event types describe how alternatives were sampled. Detector positives were defined from changes in the stable destination or target identity. Table \ref{tab:generation_timing_events} gives a more in-depth account of the events the trajectories undergo.

\begin{table}[h]
\centering
\vspace{-1.2em}
\caption{Accepted LLM call latency, executed redirection events by type, and successful trajectories by event count.}
\label{tab:generation_timing_events}
\footnotesize
\setlength{\tabcolsep}{2pt}
\renewcommand{\arraystretch}{0.88}
\begin{tabular*}{\columnwidth}{@{}l@{\extracolsep{\fill}}rrrr@{}}
\toprule
Domain & LLM calls & Mean (s) & Median (s) & 95th pct. (s) \\
\midrule
Mobile base & 167,829 & 0.229 & 0.224 & 0.313 \\
Manipulator & 123,750 & 0.092 & 0.087 & 0.116 \\
\midrule
\end{tabular*}\par\nointerlineskip
\begin{tabular*}{\columnwidth}{@{}l@{\extracolsep{\fill}}lr@{}}
Domain & Redirection event type & Count \\
\midrule
Mobile base & Same shelf/new approach & 52,800 \\
            & Nearby shelf cone       & 56,572 \\
            & Random shelf            & 58,457 \\
Manipulator & Same target/regrasp     & 21,168 \\
            & Nearby/context object   & 40,707 \\
            & Different/random object & 61,875 \\
\midrule
\end{tabular*}\par\nointerlineskip
\begin{tabular*}{\columnwidth}{@{}l@{\extracolsep{\fill}}rrrr@{}}
Domain & 0 events & 1 event & 2 events & 3+ events \\
\midrule
Mobile base & 34,000 & 26,400 & 13,200 & 26,400 \\
Manipulator & 25,500 & 12,375 & 12,375 & 24,750 \\
\bottomrule
\end{tabular*}
\vspace{-3.2em}
\end{table}

\subsection{Simulator Scenes}
\label{subsec:training_trajectory_visuals}
\begin{figure}[h]
\centering
\includegraphics[width=0.43\columnwidth]{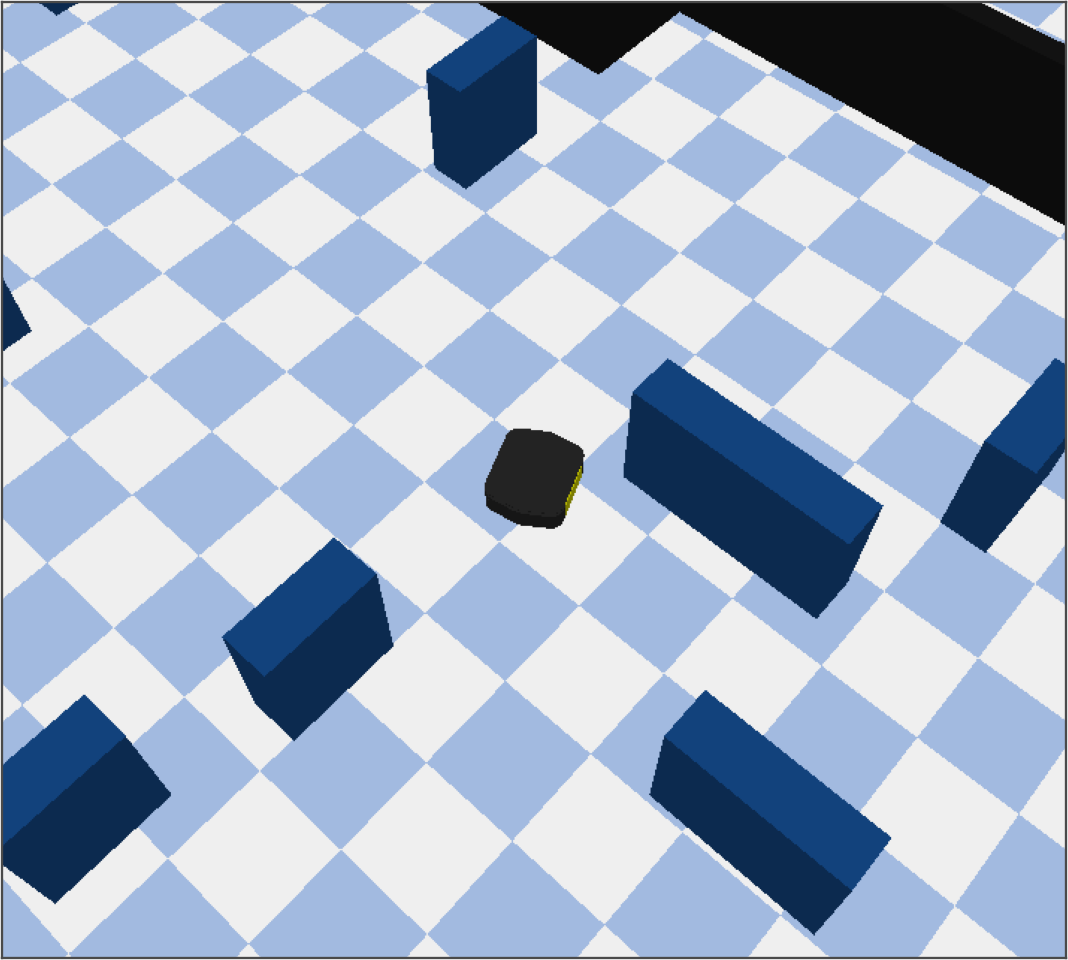}
\hspace{0.4em}
\includegraphics[width=0.43\columnwidth]{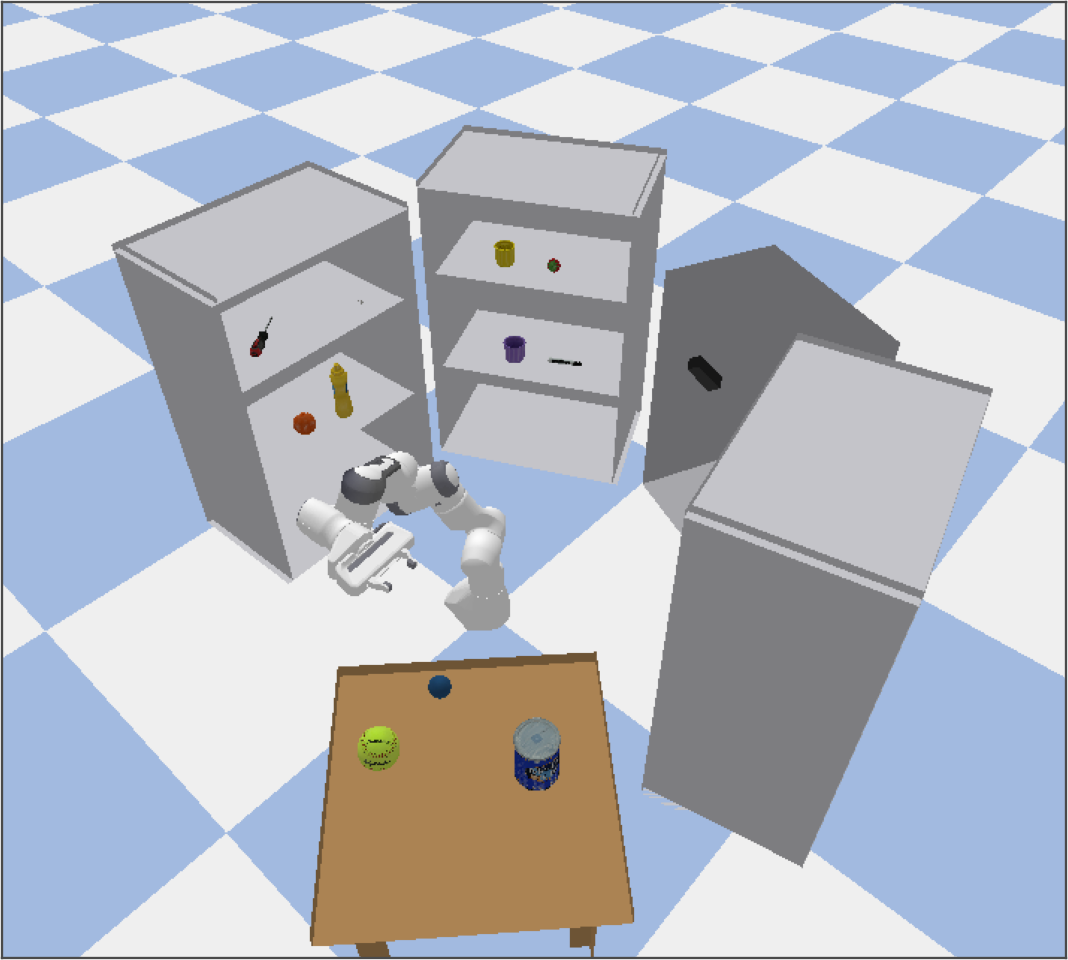}
\caption{PyBullet scenes showing Ridgeback on maps with obstacles (left) and Franka with objects, supports and grasp context (right).}
\label{fig:simulator_replays}
\vspace{-0.2em}
\end{figure}

The PyBullet simulators provide the physical environments used for dataset generation. Mobile-base scenes contain planar map geometry, obstacles, and the Ridgeback robot, while manipulation scenes contain the Franka Panda, support surfaces, objects, handles, and grasp-related geometry. Figure \ref{fig:simulator_replays} shows a representative scene from each domain.

Mobile-base scenes vary obstacle density, corridor geometry, and robot pose, while manipulation scenes vary support geometry, cabinet and table placement, and object configuration. This variation changes the physical constraints under which robot motion and redirection intent-switch events are generated. Scene geometry is used for simulation, planning, and interpretation of the resulting trials.

\section{Training and Results} \label{sec:neural_detector_results}
\subsection{Dataset Splits}
Table \ref{tab:final_dataset_splits} gives the training, validation, and testing partitions used for the final detector results. The mobile-base dataset contains 100,000 trajectories split into 70,000 training, 15,000 validation, and 15,000 testing trajectories. The selected manipulator development corpus contains 75,000 trajectories split into 52,555 training, 11,194 validation, and 11,251 testing trajectories. Splitting was performed at the trajectory level before window extraction, and normalization statistics were fitted from training data only.

\begin{table}[h]
\vspace{-1.2em}
\centering
\caption{Detector training splits for the mobile-base and manipulation datasets.}
\label{tab:final_dataset_splits}
\footnotesize
\renewcommand{\arraystretch}{0.86}
\setlength{\tabcolsep}{3pt}
\begin{tabular*}{\columnwidth}{@{\extracolsep{\fill}} l r r r r @{}}
\toprule
Quantity & Train & Validation & Test & Total \\
\midrule
\multicolumn{5}{@{}l}{\textit{Navigation}} \\
\toprule
Trajectories & 70,000 & 15,000 & 15,000 & 100,000 \\
Rows & 4,131,891 & 880,750 & 885,240 & 5,897,881 \\
\midrule
\multicolumn{5}{@{}l}{\textit{Manipulator}} \\
\toprule
Trajectories & 52,555 & 11,194 & 11,251 & 75,000 \\
Rows & 3,626,625 & 773,542 & 774,833 & 5,175,000 \\
\bottomrule
\end{tabular*}
\vspace{-1.8em}
\end{table}

The hardware configuration includes a workstation with an Intel Core Ultra 9 275HX CPU, 64 GiB RAM, an NVIDIA GeForce RTX 5080 Laptop GPU with 16,303 MiB memory, NVIDIA driver 595.71.05, and Ubuntu Linux kernel 6.17.0-35-generic. Detector training uses $\Delta t=0.05$ s and a 0.5 s post-event detection window. Training partitions fit the TCN-GRU core and postprocessor weights, and we report final metrics on the fixed test set. The final postprocessor thresholds are $\theta^b=0.76$ for navigation and $\theta^m=0.55$ for manipulation.

\subsection{Training Hyperparameters}
The navigation and manipulation detectors use the same two-stage architecture but separate trained weights and domain input tensors. Table~\ref{tab:hyperparams} summarizes the final configurations.

\begin{table}[h]
\vspace{-1.2em}
\centering
\caption{Final detector training configurations.}
\label{tab:hyperparams}
\footnotesize
\setlength{\tabcolsep}{3pt}
\renewcommand{\arraystretch}{0.88}
\begin{tabular*}{\columnwidth}{@{\extracolsep{\fill}}lcc@{}}
\toprule
Setting & Navigation & Manipulator \\
\midrule
Sampling period & 0.05 s & 0.05 s \\
Robot motion history & 3.0~s & 3.0~s \\
Latent dimension & 192 & 192 \\
Encoder & TCN + GRU & TCN + GRU \\
Chunk / stride & 64 / 8 rows & 64 / 8 rows \\
Core-training epochs & 8 & 16 \\
Loss weights \\ $(\lambda_{\mathrm{haz}},\lambda_{\mathrm{prox}},\lambda_{\mathrm{onset}},\lambda_{\mathrm{fore}})$ & 4/3/2/0.1 & 4/3/2/0.1 \\
Weight decay & $10^{-5}$ & $10^{-5}$ \\
Selected threshold & 0.76 & 0.55 \\
\bottomrule
\end{tabular*}
\vspace{-3.2em}
\end{table}

\begin{figure*}[!hb]
    \centering
    \includegraphics[width=\textwidth]{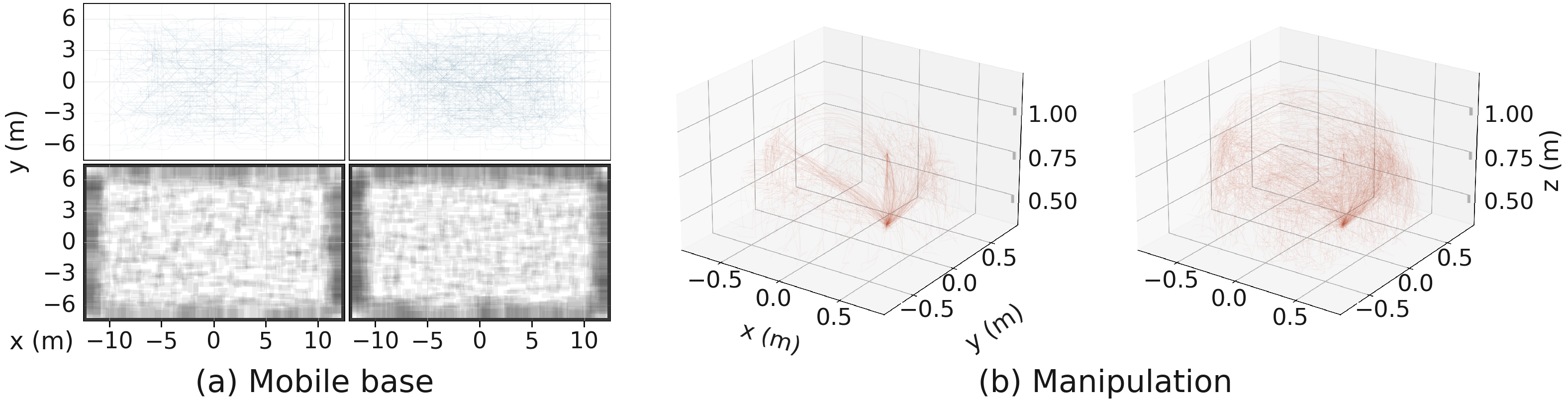}
    \vspace{-1.8em}
    \caption{(a) Mobile-base trajectories in map coordinates (top) and occupancy-map overlays from 250 trajectories per group (bottom). (b) End-effector trajectories in workspace coordinates. Each trajectory plot overlays 250 paths with no target change (left) or one or more target changes (right).}
\label{fig:training_overlay_trajectories}
\end{figure*}
\vspace{-0.6em}

\subsection{Metrics}
The primary metric was event-level switch detection within 0.5~s of occurrence. Let $\tau_q$ denote true event $q$ and $\mathcal{A}^d$ the reported alarm times in domain $d$.  Rising threshold crossings of $\pi_t^d$ at $\theta^d$ produce alarms, with at least 0.25~s between reported alarms (half the 0.5~s detection window) to avoid counting repeated detections associated with the same intent switch.

Each event is matched to the earliest unused alarm within the following 1.0~s of the same trajectory, with at most one alarm assigned to each event and each alarm assigned at most once.
Let $\ell_q^d$ be the matched alarm time minus $\tau_q$, or $+\infty$ if unmatched. For $Q_d$ true events, the fraction detected within 0.5~s of onset is
\vspace{-1em}
\begin{equation*}
R_{0.5}^{d}=\frac{1}{Q_d}\sum_{q=1}^{Q_d}
\mathbb{I}\!\left[0\le\ell_q^d<0.5\,\mathrm{s}\right].
\end{equation*}
\vspace{-1.4em}

False switches per minute (FS/min) are computed as $\mathrm{FS/min}^{d}=N_{\mathrm{false}}^{d}/(T_{\mathrm{eval}}^{d}/60)$, where $N_{\mathrm{false}}^{d}$ is the number of unmatched alarms and $T_{\mathrm{eval}}^{d}$ is the evaluated trajectory duration in seconds.

Row accuracy, precision, recall, and F1 compare thresholded probabilities with row labels, providing secondary measures of sample-level classification. Event-level detection remains the primary measure because GUIDER 2.0 uses the detector to identify discrete intent redirections. The fixed operating thresholds were $\theta^{\mathrm{base}}=0.76$ and $\theta^{\mathrm{manip}}=0.55$.

\subsection{Detector Training Results}
Figure \ref{fig:training_overlay_trajectories} illustrates trajectories with and without target changes, while Table~\ref{tab:performance} reports the final switch-detection results on the testing sets. At the validation thresholds, the navigation detector identifies 16,573 of 17,103 switch events within 0.5 s, corresponding to 96.90\% event-level detection (Wilson 95\% CI 96.63-97.15\%). The manipulation detector identifies 18,363 of 18,527 events within 0.5 s, corresponding to 99.11\% (98.97-99.24\%). 

\begin{table}[h]
\centering
\vspace{-0.5em}
\caption{Switch detection on the testing sets. Higher is better except for false switches/min.}
\label{tab:performance}
\footnotesize
\renewcommand{\arraystretch}{0.88}
\setlength{\tabcolsep}{3.5pt}
\begin{tabular*}{\columnwidth}{
@{\extracolsep{\fill}}
L{0.44\columnwidth}
R{0.23\columnwidth}
R{0.23\columnwidth}
@{}}
\toprule
Quantity & Navigation & Manipulator \\
\midrule
Events detected within 0.5~s   & 0.969 & 0.991 \\
Accuracy / F1                  & 0.828 / 0.393 & 0.997 / 0.994 \\
Sample precision / recall      & 0.606 / 0.291 & 0.995 / 0.993 \\
False switches / min  & 1.568 & 0.147 \\
\bottomrule
\end{tabular*}
\vspace{-3.5em}
\end{table}

\section{GUIDER Evaluations} \label{sec:guider_evaluation_protocol}
The GUIDER evaluation isolates the effect of switch-conditioned temporal belief updates by running each trial with both the original and the corresponding switch-aware formulations. The aim is to test whether GUIDER 2.0 preserves final target correctness and changes the time at which the correct target becomes dominant.

The simulated evaluation reuses the 100 Isaac Sim trials from the original GUIDER study \citep{contreras2025guider}, in which 20 participants each performed five tasks T1-T5 for a total of 100 trials. The tasks correspond to: T1 direct R3 cube grasp, T2 base redirection from R4 to R3, T3 manipulation redirection between cubes in R3, T4 semantic drill grasp in R1, and T5 combined base and manipulation redirection from R1 to R2 with can-to-fruit target change. Manipulation-only real-world teleoperated data is also used, as a second test \citep{kenny2026guider}.

\noindent\textbf{Evaluation Metrics.}
For trial $s$ and method $M$, let $\tau_s$ be the terminal evaluation time, $C_s$ the correct target set, and $\mathcal{Y}_{i,s}^{M}$ the predicted top set at sample time $t_i\le\tau_s$. Final correct-top is

\vspace{-1em}
\begin{equation*}
A_s^{M}=\mathbb{1}\!\left[C_s\cap\mathcal{Y}_{i_s,s}^{M}\neq\emptyset\right],
\qquad \hspace{-1em} i_s=\max\{i:t_i\le\tau_s\}
\end{equation*}
\vspace{-2.0em}

For the base, $C_s$ contains the target rectangle ID. For manipulation object mode, it contains the candidate nearest the target center, and for grasping mode, it contains all valid grasp labels associated with the target object. 

Let $h_s$ denote the sustained-correct duration used by the evaluation protocol: $h_s=0$ for base trials, $0.5$ s for Isaac manipulation trials, and $0.1$ s for the recorded real-world trials. We define
\vspace{-1em}
\begin{equation*}
\mathcal{H}_s^M=\left\{t_i\;\middle|\;
\begin{aligned}
&t_i\le\tau_s-h_s,\\
&C_s\cap\mathcal{Y}_{j,s}^{M}\neq\emptyset\quad\forall j\\
&\text{with }t_i\le t_j\le t_i+h_s
\end{aligned}\right\}.
\end{equation*}
\vspace{-1.2em}

\noindent If $\mathcal{H}_s^M\neq\emptyset$, then $t_{\mathrm{first},s}^M=\min\mathcal{H}_s^M$. For trial timing, the remaining time at confident prediction (RTCP) is

\vspace{-2.1em}
\begin{equation*}
\mathrm{RTCP}_s^M=
\begin{cases}
\tau_s-t_{\mathrm{first},s}^M, & \mathcal{H}_s^M\neq\emptyset,\\
0, & \mathcal{H}_s^M=\emptyset.
\end{cases}
\end{equation*}
\vspace{-1.5em}

Thus, a method that never reaches sustained correct-top contributes zero useful lead time, and all 100 Isaac trials remain in each paired comparison. Higher RTCP is better. In redirection trials, operators are instructed to pursue the initial target before receiving the redirection instruction. In the recorded real-world analysis, time at confident prediction (TCP) is the elapsed time from probability initialization to the first sustained correct prediction, $\mathrm{TCP}_s^M=t_{\mathrm{first},s}^M$. Because paired methods share the same terminal time, the paired improvement satisfies $\mathrm{TCP}_{\mathrm{Original}}-\mathrm{TCP}_{\mathrm{Integrated}}=\mathrm{RTCP}_{\mathrm{Integrated}}-\mathrm{RTCP}_{\mathrm{Original}}$ whenever both TCPs are defined.

The paired difference in prediction lead time is reported as $\Delta_{\mathrm{RTCP}}=\mathrm{RTCP}_{\mathrm{Integrated}}-\mathrm{RTCP}_{\mathrm{Original}}$, so positive values indicate earlier GUIDER 2.0 prediction. GUIDER 2.0 timings are categorized as earlier/same/later (E/S/L). The pooled paired Wilcoxon test is the primary pooled comparison. $p_2$ is two-sided, $p_1$ is one-sided for GUIDER 2.0 $>$Original, and $r_{\mathrm{bs}}$ is the matched-pairs rank-biserial effect size.

\begin{table*}[t]
  \centering
  \footnotesize
  \caption{Paired base-region RTCP results by task group.}
  \label{tab:base_rtcp_integrated}

  \setlength{\tabcolsep}{4.0pt}
  \renewcommand{\arraystretch}{1.10}

  \begin{tabular*}{\textwidth}{
  @{\extracolsep{\fill}}
  l c c c c c
  @{}}
    \toprule
    \textbf{Case} &
    \multicolumn{2}{c}{\textbf{RTCP median$\pm$SD [s]}} &
    \textbf{$p_2/p_1/r_{\mathrm{bs}}$} &
    \textbf{E/S/L [\%]} &
    \textbf{Mean shift [s]} \\
    \cmidrule(lr){2-3}
    &
    \textbf{Original} &
    \textbf{GUIDER 2.0} &
    &
    &
    \\
    \midrule

    T1 Direct (R3) &
    $41.00\pm10.50$ &
    $40.18\pm8.20$ &
    \textbf{.013}/.007/.73 &
    $60/30/10$ &
    $+2.38$ \\

    T2 Base redirection (R4$\rightarrow$R3) &
    $40.75\pm15.19$ &
    $39.78\pm14.92$ &
    $.579/.290/.16$ &
    $45/15/40$ &
    $+0.04$ \\

    T3 Manipulation redirection (R3) &
    $40.40\pm14.18$ &
    $41.38\pm14.12$ &
    $.635/.318/.16$ &
    $35/35/30$ &
    $+0.30$ \\

    T4 Semantic tool grasp (R1) &
    $30.34\pm10.35$ &
    $30.65\pm10.18$ &
    $.500/.250/.67$ &
    $10/85/5$ &
    $+0.18$ \\

    T5 Combined redirection (R1$\rightarrow$R2) &
    $31.32\pm8.92$ &
    $31.18\pm9.15$ &
    $.095/.958/-.50$ &
    $20/25/55$ &
    $-0.17$ \\

    \midrule

    ALL &
    $37.52\pm13.06$ &
    $38.57\pm12.81$ &
    \textbf{.032}/.016/.31 &
    $34/38/28$ &
    $+0.54$ \\

    \bottomrule
  \end{tabular*}

  \vspace{0.3em}

  \begin{minipage}{0.99\textwidth}
  \footnotesize
  RTCP is measured before contact; higher is better. Mean shift is GUIDER 2.0 $-$Original, so positive values indicate earlier GUIDER 2.0 prediction. E/S/L denotes GUIDER 2.0 earlier/same/later. $p_2$ and $p_1$ are the two- and one-sided paired Wilcoxon tests, and $r_{\mathrm{bs}}$ is the matched-pairs rank-biserial effect size. Bold $p_2$ values are significant at $p<.05$.
  \end{minipage}
\end{table*}

\subsection{Mobile-Base Evaluation}
Original GUIDER and GUIDER 2.0 are compared on the same 100 base-region trials using the same region and contact-time target definition. The comparison tests whether the switch-aware update preserves final target-region correctness and maintains or improves the RTCP. 

\begin{figure}[hb]
  \centering
  \includegraphics[width=\linewidth]{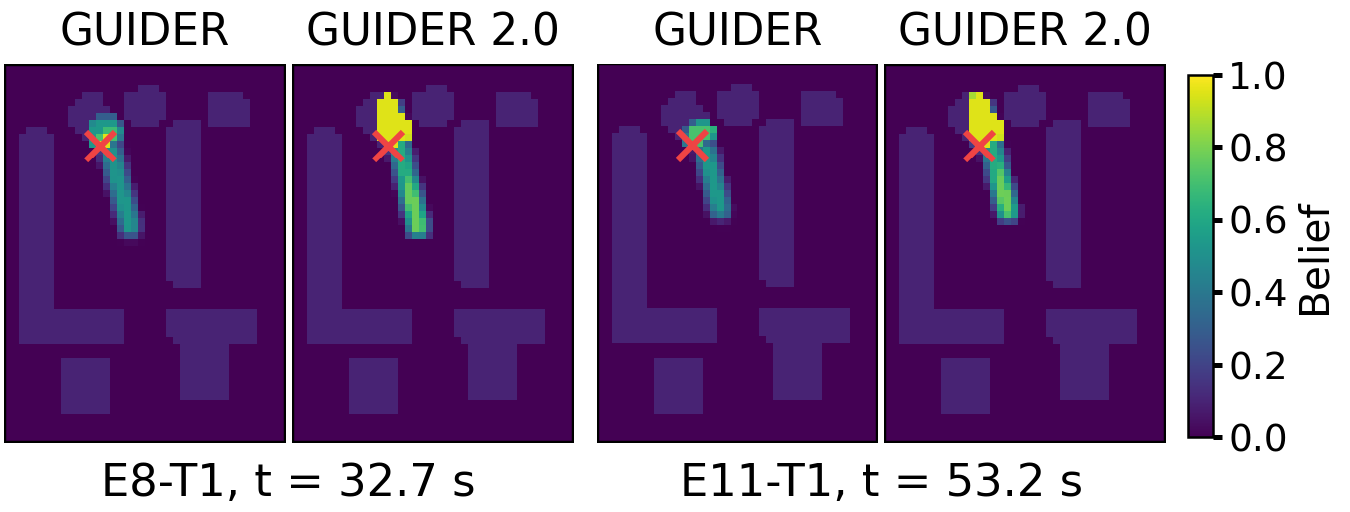}
  \vspace{-1.8em}
\caption{Mobile-base beliefs from GUIDER and GUIDER 2.0 at the same frame in each trial. The red cross marks the highest-belief cell in each panel.}
  \label{fig:base_probability_grid}
\end{figure}

Both methods retain 100/100 Top-1 target prediction correctness at contact. With switch-conditioned belief updates GUIDER 2.0 increases pooled median RTCP from $37.52$ to $38.57$ s and increases mean RTCP by $0.54$ s. GUIDER 2.0 is earlier/same/later in $34/38/28$ trials. The pooled paired comparison is significant ($p_2=.032$, $p_1=.016$, $r_{\mathrm{bs}}=.31$), although the task-group results are not uniform. T1 shows the largest mean shift (+2.38 s, $p_2=.013$), while T2, T3, T4, and T5 do not reach two-sided significance. Table \ref{tab:base_rtcp_integrated} provides a more detailed comparison of the results. Figure \ref{fig:base_probability_grid} shows a visual comparison of the two belief fields at the same trial frame for GUIDER and GUIDER 2.0.

\subsection{Manipulator Evaluation}

The manipulation evaluation compares four modes (Object GUIDER, Grasping GUIDER, Object GUIDER 2.0, and Grasping GUIDER 2.0) across the same 100 trials. Object GUIDER uses object-level candidates while Grasping GUIDER uses the same temporal update on feasible grasp-region candidates. The GUIDER 2.0 modes preserve the same perception stack and apply neural switch gating only during temporal probability evolution to both methods. The focus of the manipulation evaluation tests is to find whether switch-aware temporal updates improve the final target ranking or the time at which the correct target becomes top in intent-changing trials, without adversely affecting the non-intent-changing data.

\begin{table}[h]
\centering
\vspace{-1.5em}
\caption{Manipulator final correctness and RTCP across GUIDER modes.}
\label{tab:manip_aggregate_median}

\footnotesize
\setlength{\tabcolsep}{3pt}
\renewcommand{\arraystretch}{0.88}

\begin{tabular*}{\columnwidth}{
@{\extracolsep{\fill}}
l c c r
@{}}
\toprule
Mode & Correct & Reached & RTCP (s) \\
\midrule
Object GUIDER       & 99/100  & 99/100  & $12.28\pm6.92$ \\
Object GUIDER 2.0 & 100/100 & 100/100 & $12.89\pm6.80$ \\
Grasping GUIDER     & 99/100  & 99/100  & $12.87\pm6.83$ \\
Grasping GUIDER 2.0 & 100/100 & 100/100 & $12.99\pm6.68$ \\
\bottomrule
\end{tabular*}

\vspace{0.2em}

\parbox{0.99\columnwidth}{%
\tiny
Correct counts correct-top predictions at contact; Reached counts trials achieving sustained correct-top before contact. RTCP is reported as median$\pm$SD over all 100 trials, assigning zero when sustained correct-top is never reached. Higher RTCP indicates earlier correct prediction.
}
\vspace{-3.2em}
\end{table}

\begin{table*}[!b]
  \centering
  \scriptsize
  \caption{Paired manipulator RTCP results by task group.}
  \label{tab:manip_pairwise_rtcp_integrated}

  \setlength{\tabcolsep}{4.2pt}
  \renewcommand{\arraystretch}{1.08}

  \resizebox{\textwidth}{!}{%
  \begin{tabular}{@{}llccccc@{}}
    \toprule
    \textbf{Mode} &
    \textbf{Task} &
    \multicolumn{2}{c}{\textbf{RTCP median$\pm$SD [s]}} &
    \textbf{$p_2/p_1/r_{\mathrm{bs}}$} &
    \textbf{E/S/L [\%]} &
    \textbf{Mean shift [s]} \\
    \cmidrule(lr){3-4}
    &
    &
    \textbf{Original} &
    \textbf{GUIDER 2.0} &
    &
    &
    \\
    \midrule

    Object &
    T1 Direct, R3 cube &
    $16.13\pm3.99$ &
    $16.13\pm3.99$ &
    $1.000/.500/1.00$ &
    $5/95/0$ &
    $0.00$ \\

    Object &
    T2 Base redirect, R4$\rightarrow$R3 &
    $11.18\pm4.00$ &
    $11.20\pm4.00$ &
    $1.000/.500/1.00$ &
    $5/95/0$ &
    $0.00$ \\

    Object &
    T3 Manip. redirect, cube$\rightarrow$cube &
    $4.13\pm2.18$ &
    $4.60\pm2.40$ &
    \textbf{.008}/.004/1.00 &
    $40/60/0$ &
    $+0.68$ \\

    Object &
    T4 Semantic tool grasp, drill &
    $10.84\pm4.35$ &
    $10.84\pm4.84$ &
    $.062/.031/1.00$ &
    $25/75/0$ &
    $+0.49$ \\

    Object &
    T5 Combined redirect, can$\rightarrow$fruit &
    $20.64\pm5.28$ &
    $20.64\pm5.28$ &
    \textemdash/\textemdash/\textemdash &
    $0/100/0$ &
    $0.00$ \\

    \midrule
    Object &
    ALL &
    $12.28\pm6.92$ &
    $12.89\pm6.80$ &
    \textbf{$<.001$}/$<.001$/1.00 &
    $15/85/0$ &
    $+0.23$ \\

    \midrule

    Grasping &
    T1 Direct, R3 cube &
    $16.13\pm4.09$ &
    $16.13\pm4.09$ &
    $1.000/.500/1.00$ &
    $5/95/0$ &
    $+0.01$ \\

    Grasping &
    T2 Base redirect, R4$\rightarrow$R3 &
    $11.28\pm4.20$ &
    $11.28\pm4.19$ &
    $.500/.250/1.00$ &
    $10/90/0$ &
    $+0.01$ \\

    Grasping &
    T3 Manip. redirect, cube$\rightarrow$cube &
    $4.00\pm2.51$ &
    $4.77\pm2.54$ &
    \textbf{.004}/.002/1.00 &
    $45/55/0$ &
    $+0.69$ \\

    Grasping &
    T4 Semantic tool grasp, drill &
    $10.98\pm4.05$ &
    $10.98\pm4.35$ &
    $.062/.031/1.00$ &
    $25/75/0$ &
    $+0.30$ \\

    Grasping &
    T5 Combined redirect, can$\rightarrow$fruit &
    $20.64\pm5.18$ &
    $20.64\pm5.18$ &
    \textemdash/\textemdash/\textemdash &
    $0/100/0$ &
    $0.00$ \\

    \midrule
    Grasping &
    ALL &
    $12.87\pm6.83$ &
    $12.99\pm6.68$ &
    \textbf{$<.001$}/$<.001$/1.00 &
    $17/83/0$ &
    $+0.20$ \\

    \bottomrule
  \end{tabular}%
  }

  \vspace{0.3em}

  \begin{minipage}{0.99\textwidth}
  \tiny
  RTCP is measured before contact; higher is better. Trials without sustained correct-top are assigned RTCP$=0$, so all 100 trials contribute. Mean shift is GUIDER 2.0$-$Original; positive values indicate earlier GUIDER 2.0 prediction. E/S/L denotes GUIDER 2.0 earlier/same/later. $p_2$ and $p_1$ are two- and one-sided paired Wilcoxon tests, and $r_{\mathrm{bs}}$ is the matched-pairs rank-biserial effect size. Bold $p_2$ values are significant at $p<.05$.
  \end{minipage}
\vspace{-1.8em}
\end{table*}

\begin{figure*}[h]
  \centering
  \includegraphics[width=0.99\textwidth]{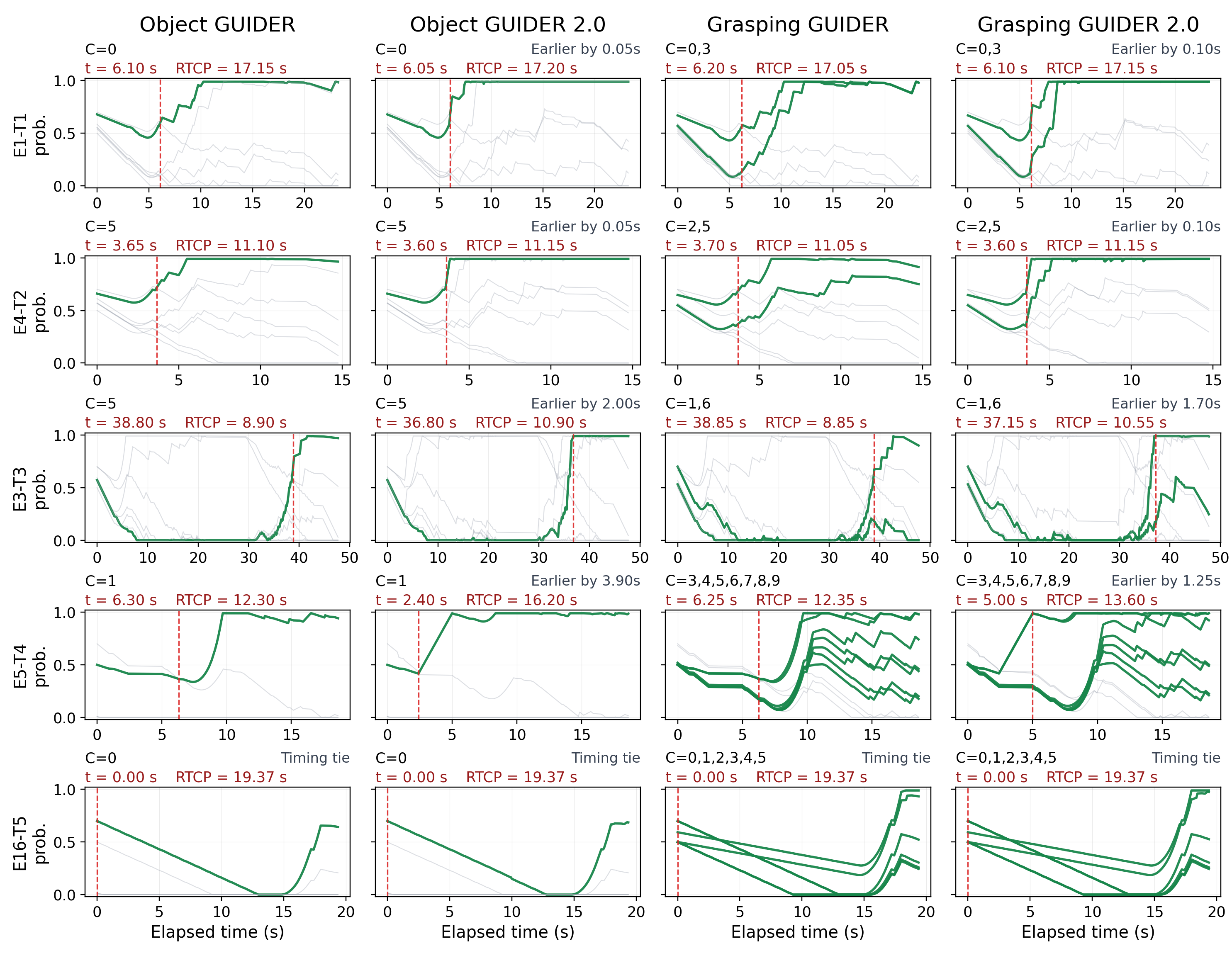}
  \caption{Probability histories for five representative trials, one per task condition. Green curves show probabilities for correct targets or grasps in each mode, and grey curves show the rest. The red dashed line marks the first sustained correct top prediction. RTCP gives the time remaining until contact.}
  \label{fig:manipulator_probability_grid}
\end{figure*}

Switch-aware manipulation improves final target ranking in the single baseline failure for each mode, with Object GUIDER and Grasping GUIDER reaching 99/100 final correct-top, whereas Object GUIDER 2.0 and Grasping GUIDER 2.0 each reach 100/100 (Table \ref{tab:manip_aggregate_median}). Under the all-trial RTCP rule, Object GUIDER 2.0 increases mean RTCP by $0.23$ s, with E/S/L $15/85/0$, $p_2<.001$, and $r_{\mathrm{bs}}=1.00$. Grasping GUIDER 2.0 increases mean RTCP by $0.20$ s, with E/S/L $17/83/0$, $p_2<.001$, and $r_{\mathrm{bs}}=1.00$. The largest task-level shifts occur in T3 (Redirection) and T4 (semantic tool different size objects) for both object and grasping modes, although only T3 reaches two-sided significance. A more detailed look at the results and comparisons can be found in Table \ref{tab:manip_pairwise_rtcp_integrated}, with a visual display of some of the earlier sustained correct predictions in Figure \ref{fig:manipulator_probability_grid}.

\newpage

\subsection{Real-world GUIDER 2.0 Grasping Demonstration} \label{subsec:realworld_integrated_grasping}
The real-world demonstration uses 20 previously recorded manipulation teleoperation episodes \citep{kenny2026guider} from three scenarios: Making Tea, Fetching Medicine, and Finding Belongings. The analysis uses the recorded robot-topic streams throughout all the experiments. Each trial is evaluated as a manipulation episode in which the operator first observes the scene, triggers GUIDER perception, and then moves from observation to action on the selected object. We treat this transition as a reference for a change in intent, as the operator is changing from reaching and positioning himself at the table (target) to going for an object (new target).

Grasping GUIDER and Grasping GUIDER 2.0 use the same perception output, candidate labels, and initial candidate probabilities. They differ only in the temporal probability evolution after candidate initialization. The GUIDER 2.0 grasping mode processes successive motion samples from the start of probability evolution. The probabilities are computed in the same order a human-facing interface would execute them. First, exploring/pausing while performing perception calculations, initializing grasping candidates, continuing the robot motion until the desired action, and updating the target probabilities until the operator's first grasp action.

\begin{figure*}[ht]
  \centering
  \includegraphics[width=0.99\textwidth]{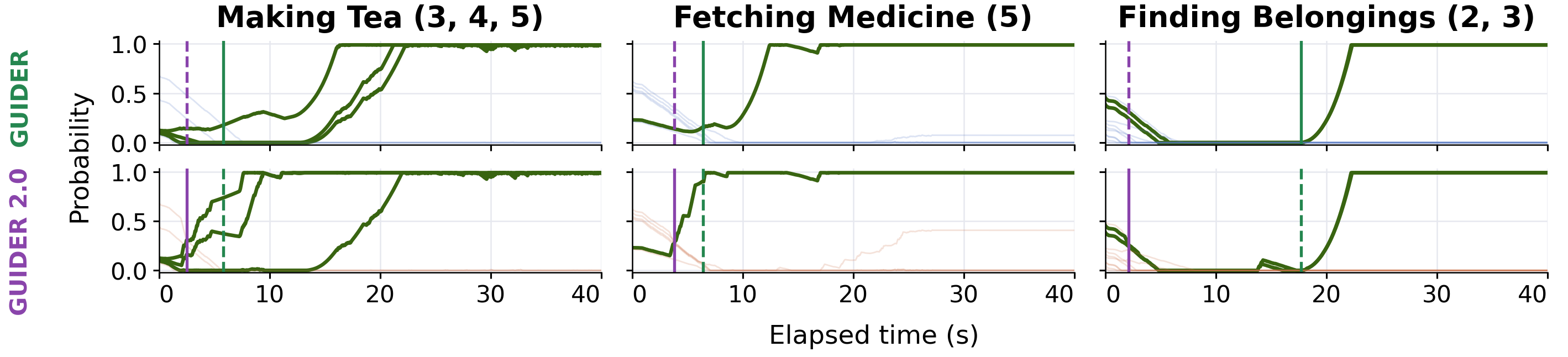}
\caption{Target probabilities in three real-world grasping episodes with shared perception initialization. Vertical lines mark TCP for GUIDER (green) and GUIDER 2.0 (purple). The solid line denotes the method shown and the dashed line the other method.}
\label{fig:realworld_probability_matrix}
\end{figure*}

\begin{figure*}[ht]
  \centering
  \includegraphics[width=0.99\textwidth]{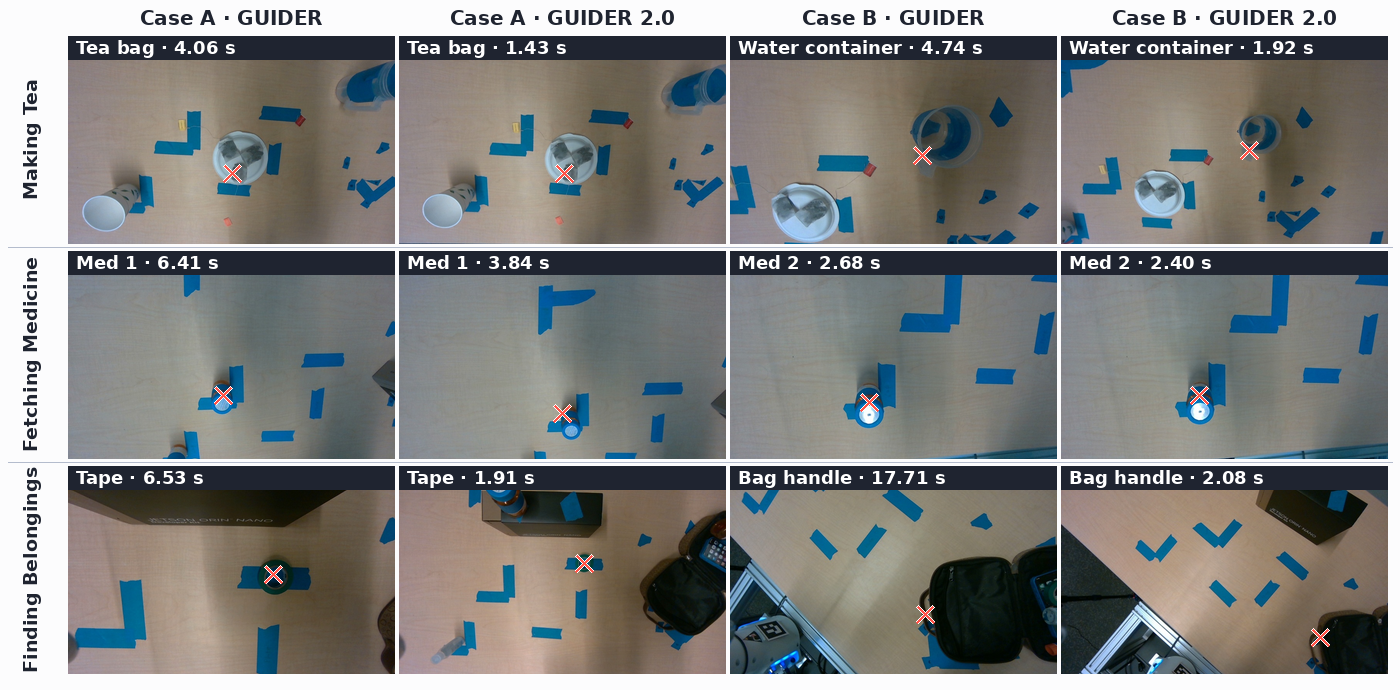}
\caption{Camera frames from two grasping episodes per scenario, shown when each method first sustained the correct target as its top prediction (TCP).}
\label{fig:realworld_tcp_camera_examples}
\end{figure*}

Across the 20 real-world episodes, Grasping GUIDER 2.0 reaches the sustained correct-top earlier in 13 cases, at the same time in seven cases, and later in no cases. Mean TCP improves by $1.73$ s, and the median paired improvement is $0.14$ s, with the paired Wilcoxon comparison being significant ($p_2<.001$, one-sided $p_1<.001$) with $r_{\mathrm{bs}}=1.00$. The effect is visible across most examples (Table \ref{tab:realworld_integrated_wilcoxon}). Figure \ref{fig:realworld_probability_matrix} shows improved TCPs on representative tests, and Figure \ref{fig:realworld_tcp_camera_examples} shows the operator views when each method first sustains the correct top prediction.

\begin{table*}[t]
\centering
\caption{Real-world grasping prediction times for GUIDER and GUIDER 2.0.}
\label{tab:realworld_integrated_wilcoxon}
\footnotesize
\setlength{\tabcolsep}{3pt}
\renewcommand{\arraystretch}{0.88}
\begin{minipage}[t]{0.49\textwidth}
  \centering
  \textbf{(a) Overall timing across 20 paired episodes}\par\smallskip
  \begin{tabular*}{\linewidth}{@{\extracolsep{\fill}}lrr@{}}
    \toprule
    Method & TCP (s) & RTCP (s) \\
    \midrule
    Grasping GUIDER     & $2.97\pm4.07$ & $41.76\pm21.36$ \\
    Grasping GUIDER 2.0 & $1.91\pm1.47$ & $43.05\pm22.59$ \\
    \midrule
    Median $\Delta$     & $+0.14$ & $+0.14$ \\
    Mean $\Delta$       & $+1.73$ & $+1.73$ \\
    \bottomrule
  \end{tabular*}\par\smallskip
  \raggedright
  Method values are median$\pm$SD. Lower TCP and higher RTCP are better. Positive $\Delta$ favours Grasping GUIDER 2.0. E/S/L$=13/7/0$, $p_2<.001$, $p_1<.001$, and $r_{\mathrm{bs}}=1.00$.\par
\end{minipage}\hfill
\begin{minipage}[t]{0.47\textwidth}
  \centering
  \textbf{(b) TCP improvement by scenario}\par\smallskip
  \begin{tabular*}{\linewidth}{@{\extracolsep{\fill}}lrr@{}}
    \toprule
    Scenario & $N$ & Mean improvement (s) \\
    \midrule
    Making Tea         & 6 & $+1.46$ \\
    Fetching Medicine  & 8 & $+0.70$ \\
    Finding Belongings & 6 & $+3.38$ \\
    \bottomrule
  \end{tabular*}\par\smallskip
  \raggedright
  Positive values indicate an earlier sustained top ranking for the correct target with Grasping GUIDER 2.0.\par
\end{minipage}

\end{table*}

\section{Discussion}
The evidence supports three main findings. First, we evaluated the switch detector on the simulated testing sets. During training data generation, the LLM used task context to select among simulator-provided targets. Navigation identifies 16,573/17,103 true switch events within 0.5 s, corresponding to 96.90\% event-level detection at a probability of 0.76 or higher, while manipulation identifies 18,363/18,527, corresponding to 99.11\% at a threshold of 0.55. Despite high detection rates in both domains, false switches were more frequent in navigation (1.568/min) than in manipulation (0.147/min). 

Second, the 100-trial Isaac Sim evaluation shows what happens when that switch evidence is integrated into GUIDER. For the mobile base, Original and GUIDER 2.0 both retain 100/100 final target-region correctness, while GUIDER 2.0 increases mean RTCP by 0.54 s overall ($p_2=.032$, $r_{\mathrm{bs}}=.31$). The largest mean RTCP gain occurs in T1, a direct grasp task, showing that predictions can also stabilise earlier without an instructed redirection.

For manipulation, Object GUIDER 2.0 and Grasping GUIDER 2.0 raise final correct-top from 99/100 to 100/100, with pooled mean RTCP shifts of +0.23 s and +0.20 s and positive matched-pairs effects ($r_{\mathrm{bs}}=1.00$ for both; both $p_2<.001$). The largest manipulation RTCP shifts occur in T3 and T4. T3 includes an instructed redirection. In T4, object size and shape may influence how the operator approaches feasible grasp regions even without an instructed redirection \footnote{Only T3 reaches two-sided significance in the task-level manipulation comparisons.}.

Third, we compared the two grasping modes on 20 recorded real-robot manipulation episodes. With identical perception initialization, Grasping GUIDER 2.0 reached a sustained correct prediction earlier in 13 episodes and at the same time in seven. Across all 20 episodes, it reached this point 1.73 s earlier on average.

Several limitations remain. The 96.90\% and 99.11\% detection rates are estimates constrained to redirection patterns represented in the training data, which could be expanded in future experiments. The Isaac Sim data is also limited to patterns from 20 participants, and the real-world set contains only 20 manipulation episodes and does not test how a human operator reacts when assistance changes online. Future work should also include teleoperation experiments and adaptive switch gates to see how changes in assistance level affect goal switch detection.

\section{Conclusion}
This paper presented the GUIDER 2.0 framework that contains bounded offline data generation and online intent switch detection. The LLM selects targets only from the candidate sets supplied by the simulator, while a two-stage TCN-GRU and MLP detector estimates redirection from recent motion and, for navigation, local map information and supplies switch evidence to GUIDER 2.0 belief updates. 

On the testing sets of the simulated datasets, the detector identifies 96.90\% of navigation switches and 99.11\% of manipulation switches within 0.5 s. In Isaac Sim, all three GUIDER 2.0 modes ranked the correct target first at contact in all 100 trials. Each increased mean prediction lead time, with pooled comparisons significant at $p_2<.05$. Across 20 recorded real-robot manipulation episodes, Grasping GUIDER 2.0 reached a sustained correct prediction 1.73~s earlier on average. 

Overall the results show that the detector reliably identifies intent switches in PyBullet trajectories with LLM-selected targets, that its evidence can improve or preserve GUIDER target-belief timing in human-operated Isaac Sim trials, and that the same switch-aware manipulation update can be applied to real-robot motions. Future work will evaluate GUIDER 2.0 in closed-loop shared-autonomy experiments and test robustness to redirection patterns outside the training distributions.

\section*{Ethics Statement}
The study used simulated and recorded teleoperation data for shared autonomy research approved by the University Ethics Committee under project ERN\_3337.

\subsection*{Author contribution}
\textbf{C.A.C.}: Conceptualisation; Methodology; Software; Formal analysis; Investigation; Writing - original draft. \textbf{M. C.}: Supervision; Resources; Writing - review \& editing; Funding acquisition. \textbf{A. R.}: Visualization; Writing - review; Funding acquisition; Supervision. \textbf{M. S.}: Industrial guidance; Resources; Writing - review \& editing. \textbf{R. S.}: Project administration; Funding acquisition; Supervision; Writing - review \& editing.

\section*{Funding}
This work was funded by the Nuclear Decommissioning Authority (NDA) and supported by the United Kingdom National Nuclear Laboratory Ltd (UKNNL). In addition, it was supported by the UK Research and Innovation (UKRI) project “REBELION" under Grant 101104241.

\section*{Conflict of Interest}
The authors declare no conflict of interest.

\bibliography{references}

\end{document}